%% file: outline.tex
\documentclass{article}

\PassOptionsToPackage{numbers,compress}{natbib}

\usepackage[preprint]{neurips_2024}

\usepackage[utf8]{inputenc} 
\usepackage[T1]{fontenc}    
\usepackage{hyperref}       
\usepackage{url}            
\usepackage{booktabs, multirow}       
\usepackage{amsfonts}       
\usepackage{nicefrac}       
\usepackage{microtype}      
\usepackage{xcolor}         
\usepackage{wrapfig} 
\usepackage{amsthm}
\usepackage{relsize}
\usepackage{mathtools}
\usepackage{graphicx}
\usepackage{pdflscape} 
\usepackage{subcaption}
\usepackage{algpseudocode,algorithm}
\usepackage{paralist}
\usepackage{amsmath, bm}
\usepackage{xcolor}
\usepackage{nccmath}
\usepackage{wrapfig}
\usepackage{tabularx}
\usepackage{upgreek}
\usepackage{pdflscape}
\usepackage{paralist}
\usepackage{subcaption}

\usepackage{hyperref}
\usepackage[nameinlink]{cleveref} 
\usepackage{enumitem}
\usepackage{fixfoot}
\usepackage{tikz}
\usepackage{longtable}
\usepackage{mdframed}

\usepackage[breakable]{tcolorbox}

\creflabelformat{equation}{#2\textup{#1}#3}

\usetikzlibrary{bayesnet}

\title{Understanding and mitigating difficulties in posterior predictive evaluation}

\author{%
  Abhinav Agrawal \\
  College of Information and Computer Science\\
  Univeristy Of Massachusetts Amherst\\
  \texttt{aagrawal@cs.umass.edu} \\
  \And
  Justin Domke \\
  College of Information and Computer Science\\
  University Of Massachusetts Amherst\\
  \texttt{domke@cs.umass.edu} \\
}
\definecolor{verylightgray}{gray}{0.9} 
\definecolor{veryverylightgray}{gray}{0.97} 

\definecolor{mycol}{rgb}{0,0,0.5}
\hypersetup{
    colorlinks,
    linkcolor={mycol},
    citecolor={mycol},
    urlcolor={mycol}
}

\newcommand{\yhwidehat}[1]{#1}
\DeclareMathOperator*{\argmax}{arg\,max}

\newcommand{\jd}[1]{{\color{red}\scriptsize}}

\newtheoremstyle{exampstyle}
  {2.5\topsep} 
  {\topsep} 
  {\itshape} 
  {} 
  {\bfseries} 
  {.} 
  {.5em} 
  {} 
  
\newtcolorbox{inferencebox}{
  breakable,
  colback=verylightgray, 
  colframe=black, 
  coltext=black, 
  boxsep=0pt, 
  arc=2mm, 
  title=,
  fonttitle=\bfseries,
  left=1mm,
  right=1mm,
  top=1mm,
  bottom=1mm,
  boxrule=0.1mm
}

\theoremstyle{exampstyle}\newtheorem{thm}{Theorem}
\theoremstyle{exampstyle}\newtheorem{defn}[thm]{Definition}
\theoremstyle{exampstyle}\newtheorem{lem}[thm]{Lemma}
\theoremstyle{exampstyle}\newtheorem{cor}[thm]{Corrolary}
\theoremstyle{exampstyle}
\theoremstyle{exampstyle}
\theoremstyle{exampstyle}\newtheorem{assump}[thm]{Assumption}
\theoremstyle{exampstyle}\newtheorem{proposition}[thm]{Proposition}
\newtheorem*{thm*}{Theorem}
\newtheorem*{claim*}{Claim}

\crefname{thm}{theorem}{theorems}
\Crefname{thm}{Theorem}{Theorems}
\crefname{lem}{lemma}{lemmas}
\Crefname{lem}{Lemma}{Lemmas}
\crefname{cor}{corollary}{corollaries}
\Crefname{cor}{Corollary}{Corollaries}
\crefname{defn}{definition}{definitions}
\Crefname{defn}{Definition}{Definitions}
\crefname{assump}{assumption}{assumptions}
\Crefname{assump}{Assumption}{Assumptions}

\global\long\def\argmax{\operatornamewithlimits{argmax}}%

\global\long\def\R{\mathbb{R}}%

\global\long\def\E{\operatornamewithlimits{\mathbb{E}}}%

\global\long\def\V{\operatornamewithlimits{\mathbb{V}}}%

\global\long\def\N{\mathcal{N}}%

\global\long\def\tr{\operatorname{tr}}%

\global\long\def\pars#1{\left(#1\right)}%

\global\long\def\pp#1{(#1)}%

\global\long\def\bracs#1{\left[#1\right]}%

\global\long\def\verts#1{\left\vert #1\right\vert }%

\global\long\def\KL#1{[#1]}%

\global\long\def\KL#1#2{\textrm{KL}\pars{#1\ \middle\Vert\ #2}}%

\newcommand{\PPDp}{\textrm{PPD}}
\newcommand{\PPDq}{\textrm{PPD}_q}
\newcommand{\SNR}[1]{\textrm{SNR}\left(#1\right)}
\newcommand{\opt}[1]{#1^{\mathrm{Opt}}}
\newcommand{\ropt}{\opt{r}}

\newcommand{\xid}{\xi_{\mathcal{D}}}

\newcommand{\xidtwodstar}{\xi_{\mathcal{D} + 2\mathcal{D^*}}}
\newcommand{\traindata}{\mathcal{D}}
\newcommand{\testdata}{\mathcal{D^*}}
\newcommand{\IWELBO}{\textrm{IW-ELBO}}

\newcommand{\genposparam}[1]{{#1}}
\newcommand{\genposconst}{\exp V}
\newcommand{\genposlogconst}{V}
\newcommand{\genpostdist}{p}

\newcommand{\genvarlogconst}{Z_\traindata}

\newcommand{\varparamone}{\eta}
\newcommand{\cteststatthree}[3]{#1(#2#3)}
\newcommand{\cteststattwo}[3]{
  #1\begin{bmatrix}
      #2(#3)\\
      |#3|
    \end{bmatrix}
}
\newcommand{\exampletraindata}{{\color{mypurple} \mathcal{D}}}
\newcommand{\examplexid}{{\color{mypurple}\xi_{\traindata}}}

\newcommand{\testdataone}{{\color{red} \mathcal{D}_{1}^{*}}}
\newcommand{\examplexidddone}{{\color{red} \xi_{\traindata + 2\testdataone}}}
\newcommand{\examplexiddone}{{\color{red} \xi_{\traindata + \testdataone}}}

\newcommand{\testdatatwo}{{\color{blue}\mathcal{D}_{2}^{*}}}

\newcommand{\examplexiddtwo}{{\color{blue}\xi_{\traindata + \testdatatwo}}}

\newcommand{\customsmaller}[2][0.85]{\scalebox{#1}{$#2$}}

\newcounter{relctr} 
\definecolor{mypurple}{rgb}{0.5019607843137255, 0.0, 0.5019607843137255}

\AtBeginDocument{} 

\allowdisplaybreaks

\newtcolorbox{examplebox}[1][]{%
    colback=veryverylightgray,
    colframe=verylightgray,
    coltitle=black,
    title=#1,
    boxrule=0.5pt,
    boxsep=2pt,
    left=5pt,
    right=5pt,
    top=5pt,
    bottom=5pt,
    fonttitle=\bfseries,
    breakable,
}

\begin{document}

\maketitle
\begin{abstract}
  Predictive posterior densities (PPDs) are of interest in approximate Bayesian inference. 
  Typically these are estimated by simple Monte Carlo (MC) averages using samples from the approximate posterior. 
  We observe that the signal-to-noise ratio (SNR) of such estimators can be extremely low.
  An analysis for exact inference reveals  SNR decays exponentially as there is increase in (a) the mismatch between training and test data, (b) the dimensionality of the latent space, or (c) the size of the test data relative to the training data.
  Further analysis extends these results to approximate inference.
  To remedy the low SNR problem, we propose replacing simple MC sampling with importance sampling using a proposal distribution optimized at test time on a variational proxy for the SNR, and demonstrate that this yields greatly improved estimates.
\end{abstract}
\section{Introduction}


A common task in approximate Bayesian inference is to calculate predictive posterior estimates. 
Given a model with prior $p(z)$ and likelihood $p(\traindata \vert z)$, an approximate inference method provides a tractable distribution $q_{\traindata}(z)$ to be used in place of the intractable posterior $p(z \vert \traindata)$  \citep{ranganath14,kucukelbir2017automatic,blei2017variational}.
The predictive posterior density (PPD) of another data set $\testdata$ under $q_\traindata$ is defined as
\begin{align}
  \PPDp \coloneqq \int p(\testdata \vert z)q_\traindata(z)dz. \label{eq: PPDq}
\end{align}
PPD is extensively used across machine learning for model selection, comparison, and criticism \citep{gelman1996posterior,gelman2013bayesian,van2021bayesian}, and making predictions and forecasts \citep{esteva2017dermatologist,filos2019systematic,kompa2021second,immer2021improving,PETROPOULOS2022705,Tyralis2022ARO,MARTIN2024811}. 
Another common use is in evaluating inference methods where higher $\PPDp$ values indicate better inference  \citep{wu2016quantitative,wenzel2020good,izmailov2021bayesian,dhaka2021challenges,agrawal2021amortized,kim2022markov,reichelt2022rethinking,zimmermann2023a,sendera2024diffusion}.
The integral in \cref{eq: PPDq} is typically intractable and is estimated via the simple Monte Carlo estimator
\begin{align} 
  R_{K} = \frac{1}{K} \sum_{k=1}^{K}p(\testdata \vert z_k), \quad \text{ where }\quad  z_1, \dots, z_K \sim q_{\traindata}(z). \label{eq: naive mc estimator for ppdq.}
\end{align}
It is common to work in log-space and estimate $\log \PPDp$ by $\log R_K$. By Jensen's inequality, the mean of $\log R_K$ estimator depends on the signal-to-noise ratio (SNR) of $R_K$. 
In this paper, we observe that the estimator in \cref{eq: naive mc estimator for ppdq.} can sometimes have extremely low SNR.
We identify three quantities that influence this: the degree of ``mismatch'' between $\testdata$ and $\traindata$, the dimensionality of $z$, and the size of $\testdata$ relative to $\traindata$. 

\begin{figure}
  \centering
  \begin{subfigure}{0.4\textwidth}
    \centering
    \includegraphics[trim = 8 10 8 2, clip, width=0.95\linewidth]{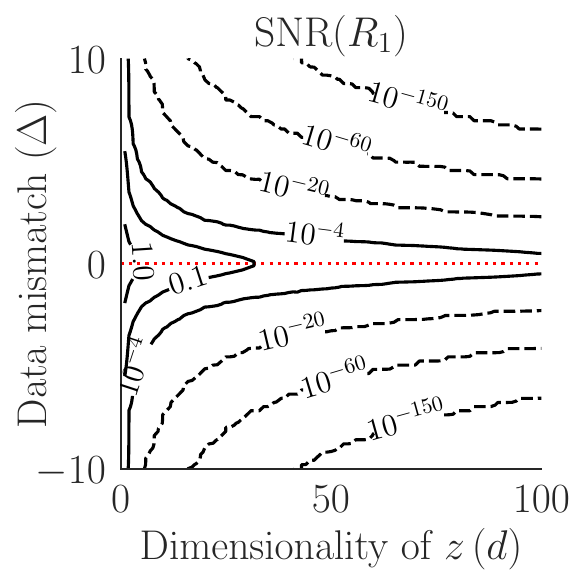}
  \end{subfigure}
  \begin{subfigure}{0.58\textwidth}
    \centering
    \includegraphics[trim = 8 10 8 2, clip, width=0.9\linewidth]{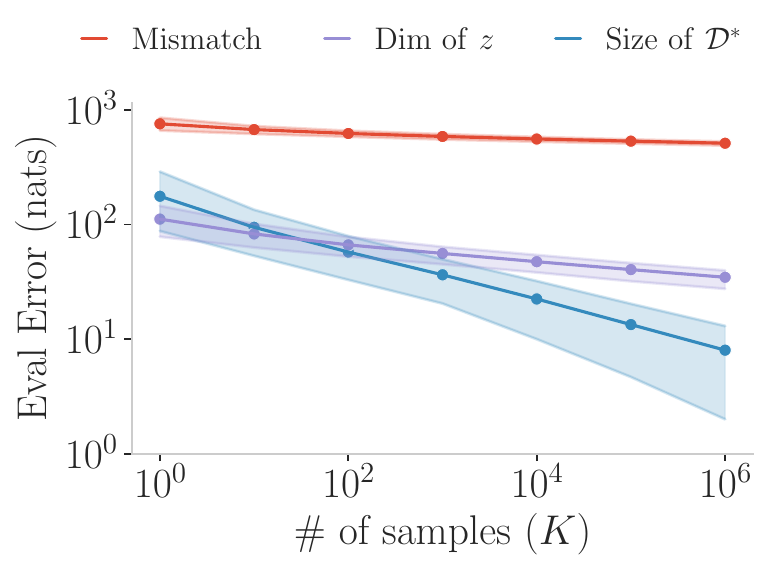}
  \end{subfigure}
\caption{
  \small{
    \textbf{Left.} 
    SNR contours of the naive MC estimator for a linear regression model when sampling from the true posterior.
    \textbf{Right.} The evaluation error, given by $\log \PPDp - \log R_K$, for the linear regression model when either data mismatch, dimensionality of $z$, or size of $\testdata$ relative to $\traindata$ is high.  
    Error is extremely poor and sometimes does not improve much with more samples.
    What explains this? How can we do better evaluation?
    \label{fig: illustration}
    }}
    \vspace{-10pt}
\end{figure}
As an example, consider a Gaussian regression model $p(z,y\vert x)$ where $y$ is the response variable, $x$ is the feature vector of $d$ dimensions, and $z$ is the regression weight vector (see \Cref{subsec: linear regression.} for more details of the model.) 
We sample a training dataset $\traindata$ and then create a test dataset $\testdata$ by adding a ``mismatch'' $\Delta$ to each response vector $y$ in $\traindata$. 
For simplicity, assume exact inference with $q_{\traindata}(z) = p(z \vert \traindata)$. As we show later, one can compute signal-to-noise ratio (SNR) for this model in closed-form (\Cref{thm: snr monte carlo gen model form.}.)
\Cref{fig: illustration}, in the left subplot, shows how quickly SNR decays as we vary mismatch $\Delta$ between $\testdata$ and $\traindata$ and dimensionality $d$ of $z$. The right subplot of \cref{fig: illustration}, shows how the mean evaluation error $\log \mathrm{PPD} - \log R_K$ varies when we independently increase the three factors influencing SNR. Evaluation errors are large and simply using more samples does not suffice.

The linear regression example hints that low SNR problems occur easily and raises two major questions: First, when will SNR of the estimator in \cref{eq: naive mc estimator for ppdq.} be high or low? Second, is there anything we can do to mitigate low SNR in practice? 
Answering these questions is the main goal of this paper.

Our first contribution (\Cref{sec: conjugate model analysis}) is to analyze the SNR problem when inference in exact. 
\Cref{thm: snr monte carlo gen model form.} provides two equivalent forms of the SNR in any model\textemdash one in terms of how much the posterior changes (measured by KL-divergence) if $\testdata$ is added to training data, and one in terms of the nonlinearity of the log-normalization constant of the posterior distribution. \Cref{cor: snr monte carlo exp family form.} extends this analysis with conjugate models, supporting the view that SNR decays exponentially in (1) data mismatch, (2) dimensionality, or (3) the size of $\testdata$ relative to $\traindata$.

Our second contribution (\Cref{sec: snr problems with approximate inference.}) is to generalize the above analysis to approximate inference.
\Cref{thm: snr monte carlo under q.} provides expressions for SNR in any model using approximate inference, and 
\Cref{cor: snr monte carlo under q and conjugacy.} provides SNR for conjugate models. 
Both support the idea that when the approximation is good, SNR decays exponentially as the three factors, mentioned earlier, increase.

Our final contribution (\Cref{sec: learned importance sampling.}) is to mitigate the SNR problem.
We propose replacing the naive MC with importance sampling (IS) where we learn a parameterized proposal $r(z)$ at test time.
We notice that SNR of the IS estimator is asymptotically related to tightness of an \emph{importance weighted} evidence lower-bound \citep{burda2015importance,maddison2017filtering,dieng2017variational,rainforth_tighter_2018,domke2018importance}, and can be optimized using standard techniques \citep{ranganath14,kucukelbir2017automatic,aagrawal2020,ambrogioni2021automatic,ambrogioni2021automaticb,burroni2023sample} to learn $r(z)$ (\Cref{fig: pseudocodes}.)
Our adaptive strategy provides vast improvements on wide range of scenarios (\Cref{sec:experiments}.) 
On a hierarchical model using MovieLens-25M \cite{harper2015movielens}, it improves performance estimates between two competing approximate inference methods by almost five-folds (\Cref{subsec: movielens 25m.}.)

\section{Analysis with exact inference}

\jd{Do we need to mention somewhere a modeling assumption, e.g. that $p(z,D)=p(z)\prod_{d \in D}p(d|z)$? Maybe only for Prop. 2?}

\label{sec: conjugate model analysis}
This section considers the PPD evaluation when inference is exact, so $q_{\traindata}(z) = p(z \vert \traindata)$.
The following result gives two equivalent forms for $\SNR{R_K}=\customsmaller[0.9]{\E[R_K]/\sqrt{\V[R_K]}}$, for any model.
(Note: We use multiset notation for datasets, so $\traindata + \testdata$ is the multiset addition and $2\traindata = \traindata + \traindata$ \citep{costa2021introduction}.)
\begin{thm}
  \label{thm: snr monte carlo gen model form.}
  Let $R_K$ be the Monte Carlo estimator for the $\PPDp$ (\cref{eq: naive mc estimator for ppdq.}) with exact inference. Let $p(z, \traindata) = p(z)\customsmaller[0.85]{\prod_{y \in \traindata}}p(y \vert z)$.
  Then, 
  $\SNR{R_K} = \sqrt{K}/\sqrt{\exp(\delta)^2 - 1}$ for
  \begin{align}
    \delta 
    & = \frac{1}{2} \KL{\genpostdist(z \vert \genposparam{\traindata + \testdata})}{\genpostdist(z \vert \genposparam{\traindata})} + \frac{1}{2} \KL{\genpostdist(z \vert \genposparam{\traindata + \testdata})}{\genpostdist(z \vert \genposparam{\traindata + 2\testdata})} \label{eq: new delta form 1.} \\
    & = \frac{\genposlogconst(\genposparam{\traindata }) + \genposlogconst(\genposparam{\traindata + 2\testdata})}{2}  - \genposlogconst(\genposparam{\traindata + \testdata}) \label{eq: new delta form 2.}
  \end{align}
  where $\genposlogconst$ is the log-normalization function $\genposlogconst(\genposparam{\traindata}) = \log \int p(\traindata \vert z) p(z) dz$. 
\end{thm}
\begin{wrapfigure}[10]{r}{0.35\textwidth}
  \vspace{-16pt}
  \centering
  \includegraphics[trim = 6 8 2 2, clip, width = \linewidth]{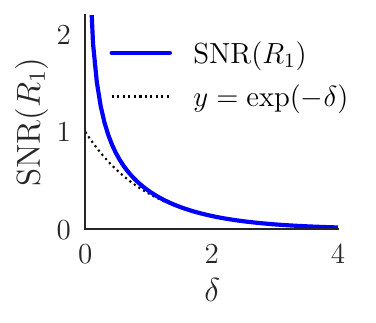}
  \caption{
  \small{SNR rapidly decays with $\delta$.}\label{fig: log snr vs delta and snr vs delta}}
\end{wrapfigure}
We provide a proof sketch and formal proof in \Cref{app: fina sec: theorem exact inference.}. To understand this result, note that if $\delta$ is reasonably large, then $\SNR{R_K} \approx \sqrt{K} \exp(-\delta)$ (\Cref{fig: log snr vs delta and snr vs delta}).

  
The KL-divergence representation in \cref{eq: new delta form 1.} shows that  SNR is determined by how different the posterior $p(z \vert \traindata+\testdata)$ is from the posteriors $p(z \vert \traindata)$ and $p(z \vert \traindata+2\testdata)$. 
Intuitively, if adding or subtracting $\testdata$ significantly changes the posterior $p(z \vert \traindata + \testdata)$, then the SNR will be small.


Now, when would adding $\testdata$ \emph{not} have significant effect on $p(z \vert \traindata + \testdata)$, meaning the SNR would be high? Intuitively, this is expected if the following three conditions are all true:
\begin{enumerate}[labelsep=0.5em,leftmargin=1.5em,partopsep=0em,topsep=0em,itemsep=0em]
\item The dataset $\traindata$ is large, so $p(z \vert \traindata)$ is concentrated.
\item The dataset $\testdata$ is similar to $\traindata$, so $p(z \vert \traindata+\testdata)$ and $p(z \vert \traindata+2\testdata)$ concentrate similar to $p(z \vert \traindata)$.
\item The dataset $\testdata$ isn't too large relative to $\traindata$, so that the posteriors involving $\testdata$ aren't much more concentrated than $p(z \vert \traindata)$.
\end{enumerate}

But even if the above conditions are satisfied, KL divergences won't be exactly zero. In \Cref{app: final sec: prop.}, we we analyze these conditions by approximating the posteriors using the Bayesian central limit theorem (CLT). The resutls of this analysis can be summarized as follows.

\begin{proposition}[Informal]
  \label{prop: delta Bayesian CLT}
Suppose $\testdata$ and $\traindata$ are large enough that posteriors in \cref{eq: new delta form 1.} are well-approximated via the Bayesian CLT as Gaussians centered at their maximum-likelihood estimates (MLEs). Also suppose that $\traindata$, $\traindata + \testdata$, and $\traindata + 2\testdata$ are similar enough that the MLE and Hessian of the \emph{average} log-likelihood is the same for all three. If $d$ is the number of dimensions of $z$, then
\begin{align}
\delta \approx \frac{d}{2} \log \frac{1 + \verts{\testdata}/\verts{\traindata}}{\sqrt{1 + 2 \verts{\testdata} / \verts{\traindata}}}.\label{eq: delta CLT}
\end{align}
\end{proposition}

Intuitively, this result says that even when the datasets are similar, $\delta$ increases linearly in the number of dimensions. It also increases in terms of the ratio $\verts{\testdata}/\verts{\traindata}$, but slowly (note that the right hand of \cref{eq: delta CLT} is well approximated by $\frac{d}{4} \log |\testdata|/|\traindata|$ when $|\testdata|/|\traindata|$ is large.) \jd{please double check this calculation}

For arbitrary datasets, there is no reason for the divergences in \cref{eq: new delta form 1.} to be small: If $\testdata$ and $\traindata$ have mismatch, then the larger the datasets are, the more the three posteriors in \cref{eq: new delta form 1.} will concentrate around different points, yielding large divergences. 
However, if $\testdata$ and $\traindata$ are similar and $\verts \traindata$ is large, then $\delta$ depends only on the number of dimensions (linearly) and ratio $\verts{\testdata}/\verts{\traindata}$ (logarithmically).

\subsection{Analysis with exact inference and conjugacy}
\label{sec: analysis with exact inference and conjugacy}
For additional insight, this section examines \Cref{thm: snr monte carlo gen model form.} in the context of conjugate models.
Consider an exponential family
\begin{align}
    p(y \vert z) = h(y)\exp(T(y)^\top \phi(z) - A(z)), \quad \text{where } \quad A(z) = \log \int h(y)\exp(T(y)^\top \phi(z)) dy, \label{eq: conjugate likelihood}
  \end{align}
$h(y)$ is the base measure, $T(y)$ is the sufficient statistic, $\phi$ is a one-to-one parameter map, and A is the log-partition function ensuring normalization.
The corresponding conjugate family is
\begin{align}
    s(z \vert \xi) = \exp\left( \xi^\top\begin{bmatrix}
        \phi(z) \\ -A(z)
    \end{bmatrix} - B(\xi)\right), \quad \text{where } \quad B(\xi) = \log \int \exp\left( \xi^\top\begin{bmatrix}
          \phi(z) \\ -A(z)
      \end{bmatrix}\right) dz. \label{ eq: normalization constant for conjugate prior}
\end{align}

This is called "conjugate" because if the prior is $p(z) = s(z \vert \xi_0)$ and the likelihood is $p(\traindata \vert z)=\prod_{y \in \traindata} p(y \vert z)$, then the posterior is within the same family and given by
\begin{align}
      p(z \vert \traindata) = s(z \vert \xid), \quad \text{where } \quad \xi_\traindata
      &
      = \xi_0 + \begin{bmatrix}
        \sum_{y \in \traindata} T(y) \\ |\traindata| 
      \end{bmatrix}.
    \label{eq: xiD}
    \end{align}
\begin{cor}
  \label{cor: snr monte carlo exp family form.}
  Take a model with likelihood $p(\traindata \vert z)$ in an exponential family (\cref{eq: conjugate likelihood}) with prior $p(z)=s(z \vert \xi_0)$ in the corresponding conjugate family (\cref{ eq: normalization constant for conjugate prior}). 
  Let $R_K$ be the Monte Carlo estimator for the PPD (\cref{eq: naive mc estimator for ppdq.}) with exact inference. Then, 
  $\SNR{R_K} = \sqrt{K}/\sqrt{\exp(\delta)^2 - 1}$ for
  \begin{align}
  \delta 
    &= \frac{1}{2}\KL{s\pp{z\vert \xi_{\traindata+\testdata}}}{s\pp{z\vert \xi_\traindata}}+\frac{1}{2}\KL{s\pp{z\vert \xi_{\traindata+\testdata}}}{s\pp{z\vert \xi_{\traindata+2\testdata}}} \label{eq: exp family delta form 1.} \\
    &= \frac{B\left(\xi_\traindata\right) + B(\xi_{\traindata + 2\testdata})}{2}  - B\left(\xi_{\traindata + \testdata}\right),\label{eq: exp family delta form 2.}
  \end{align}
  where for any dataset $\traindata$, $\xi_\traindata$ is as in \cref{eq: xiD} and $B$ is as in \cref{ eq: normalization constant for conjugate prior}.
\end{cor}

\jd{The rest of this section is heavily rewritten—please read carefully}

This result is very similar to \cref{thm: snr monte carlo gen model form.}.
The main advantage of this new result is that the second form for $\delta$ in terms of log partition functions (\cref{eq: exp family delta form 2.}) \emph{does} allows additional insight over the corresponding earlier result (\cref{eq: new delta form 2.}).
This happens because (1) $\xi_\traindata$ has a very simple relationship to $\traindata$ (\cref{eq: xiD}) and (2) $B$ is a log-partition function, and therefore has a predictable geometry. \jd{I felt a new paragraph was needed here. Good? Too short? Too long? I wrote some additional explanations but commented out above. Let me know if you think needed.}

To understand \cref{eq: exp family delta form 2.}, note that $\xi_{\traindata+\testdata}=\frac{1}{2}(\xi_\traindata + \xi_{\traindata+2\testdata}).$ Since $B$ is convex, $\delta$ is the \emph{looseness in Jensen's inequality}:  the mean of $B(\xi_\traindata)$ and $B(\xi_{\traindata+2\testdata})$ versus $B$ applied to the mean of $\xi_\traindata$ and $\xi_{\traindata+2\testdata}$.
But Jensen's inequality is tight when the function is nearly linear in the range evaluated.
Now, imagine evaluating $B(a\xi)$ for $a>0$, i.e. along a ray emanating from the origin. $B$ has a "log-sum-exp" form \citep{boyd2004convex}, so as $a$ becomes large, $B(a\xi)$ becomes nearly linear along that ray. So, when $\xid$ and $\xidtwodstar$ are large and lie near a ray emanating from the origin, $\delta$ will be small.

Thus, $\delta$ will be small (and the SNR large) when:
\begin{enumerate}[labelsep=0.5em,leftmargin=2.5em,partopsep=0em,topsep=0em,itemsep=0em]
    \item $\xi_\traindata$ is large (so that $B$ is locally "flat" near $\xi_\traindata$). \jd{I'd like to say that $\traindata$ is large, but I think I can't? Because $\xi_\traindata$ doesn't necessarily increase for more data since $T$ could cancel out?}
    \item Sufficient statistics $T(\traindata)$ and $T(\testdata)$ are similar and the prior parameters $\xi_0$ are either small or nearly proportional to $\xi_\traindata$
    (so $\xid$ and $\xidtwodstar$ lie close to a ray emanating from origin.)
  \end{enumerate}


\begin{examplebox}[Example]

Take a model with prior $p(z) = \N(z \vert 0, 1)$ and likelihood $p(y \vert z)=\N(y \vert z, \sigma^2)$, with known variance $\sigma^2$. 
Let $\customsmaller[0.9]{\overline{T}(\traindata)}$ denote the mean sufficient statistics $T(y)$ over $y\in\traindata$. 
Take a training dataset $\exampletraindata$ with $\vert\exampletraindata\vert=100$, and $\customsmaller[0.9]{\overline{T}(\exampletraindata) = 10}$ and a similar test dataset with 
$\vert\testdataone\vert=100$, and $\customsmaller[0.9]{\overline{T}(\testdataone) = 10}$.

\vspace{5pt}

\begin{wrapfigure}[14]{r}{0.72\textwidth}
  \vspace{-10pt}
  \centering
  \includegraphics[trim = 5 5 350 5,clip, width = \linewidth]{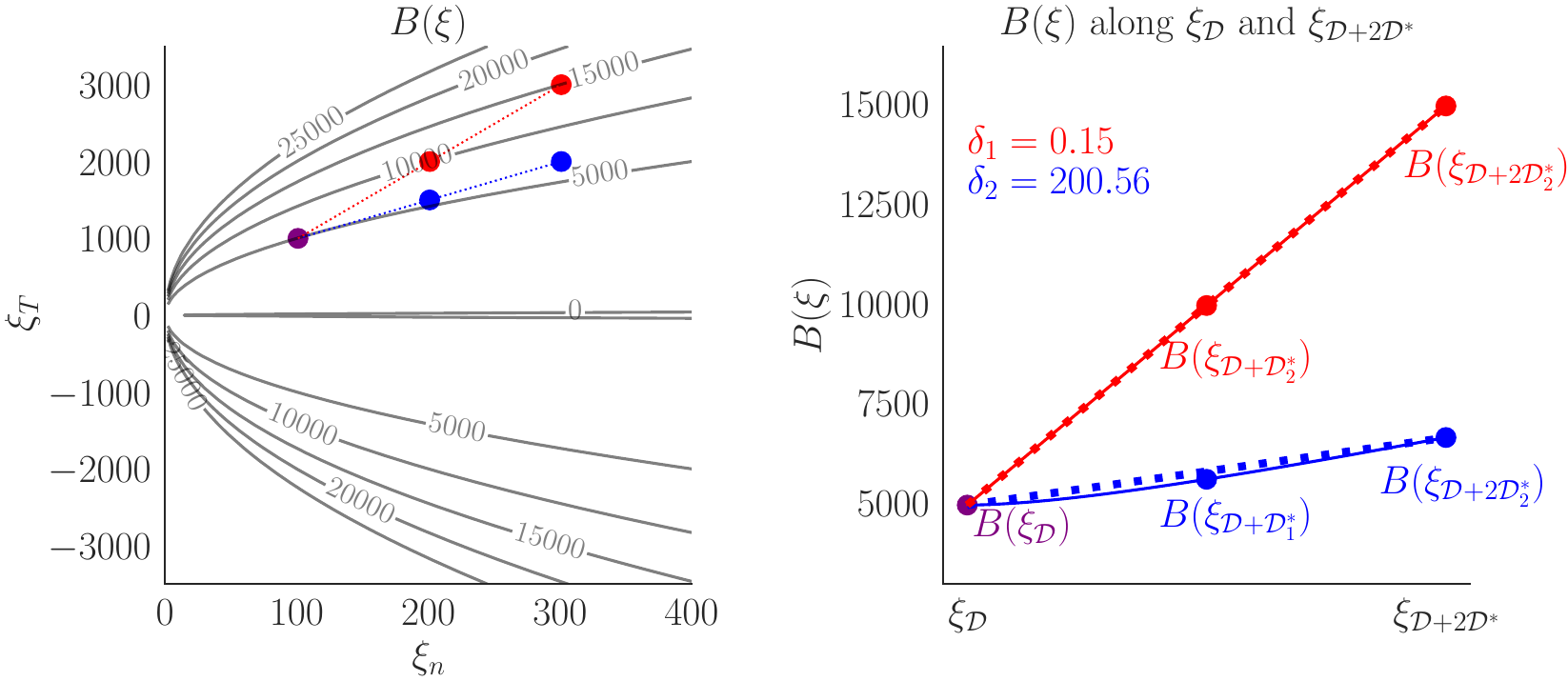}
  \caption{
    \small{
      \textbf{Left}: The log partition function $B(\xi)$ (\cref{ eq: normalization constant for conjugate prior}).
      \textbf{Right.} The values of $B(\xi)$ along the lines joining $\examplexid$ to $\examplexiddone$ and $\examplexiddtwo$. 
      } 
      \label{fig: grad b and b} 
      }
    \end{wrapfigure}
\Cref{fig: grad b and b} shows the function $B(\xi)$ along with the values of $\xi$ corresponding to each dataset.
Notice how $\examplexid, \examplexiddone,$ and $\examplexidddone$ are equidistant on a "ray" pointing to near
  the origin (left panel) meaning Jensen's inequality is nearly tight (right panel).

\vspace{5pt}

Now, take a ``mismatched'' test dataset $\testdatatwo$ with $\vert\testdatatwo\vert=100$, and $\customsmaller[0.9]{\overline{T}(\testdatatwo) = 5}$. The line joining $\examplexid$ and $ \examplexiddtwo$ does \emph{not} point towards the origin, meaning Jensen's inequality is not tight, resulting in an astronomically small SNR of
$\SNR{R_1} \approx 7.9 \times 10^{-88}$.

\end{examplebox}


\section{Analysis with approximate inference}
\label{sec: snr problems with approximate inference.}
This section generalizes the SNR analysis to approximate inference where $q_{\traindata}$ may not be the same as the true posterior. We start by generalizing \cref{eq: naive mc estimator for ppdq.}.

\begin{thm}
  \label{thm: snr monte carlo under q.}
  Let $R_{K}$ be the Monte Carlo estimator for the $\PPDp$ (\cref{eq: naive mc estimator for ppdq.}.) Then,
  $\SNR{R_{K}} = \sqrt{K}/{\sqrt{\exp(\delta)^2 - 1}}$ for
  \begin{align}
    \delta 
    & = \frac{1}{2} \KL{q_{\traindata}(z \vert \testdata)}{q_{\traindata}(z)} + \frac{1}{2} \KL{q_{\traindata}(z \vert \testdata)}{q_{\traindata}(z \vert 2\testdata)}\label{eq: delta q form 1} \\
    & = \frac{1}{2}\genvarlogconst(2\testdata)  - \genvarlogconst(\testdata) \label{eq: delta q form 2},
  \end{align}
  where $\genvarlogconst(\testdata) = \log \int p(\testdata \vert z) q_{\traindata}(z) dz$  and $q_{\traindata}( z \vert \testdata) \propto p(\testdata \vert z) q_{\traindata}(z)$.
\end{thm}
We provide a proof in \Cref{app: final sec: theorem approximate inference.}. As in previous results, \cref{eq: delta q form 1} determines $\delta$ in terms of divergences, but the distributions involved are new. One can think of $q_\traindata(z \vert \testdata)$ as the posterior that results from taking $q_\traindata(z)$ as a prior and then conditioning on $\testdata$. When inference is exact, $q_\traindata(z \vert \testdata) = p(z \vert \traindata + \testdata)$ and \cref{eq: delta q form 1} reduces to \cref{eq: new delta form 1.}. So, when inference is accurate, SNR depends on the same three factors as before: the mismatch between $\testdata$ and $\traindata$, size of the latent space $d$, and the size of $\testdata$ relative to $\traindata$. More generally, this result says that the SNR depends on how much the ``posterior'' $q_\traindata(z \vert \testdata)$ varies from the ``prior'' $q_\traindata(z)$.

\Cref{eq: delta q form 2} is also a generalization of \cref{eq: new delta form 2.}. To see this, write $\delta = \customsmaller[0.9]{\frac{1}{2} \left(\genvarlogconst(\emptyset) + \genvarlogconst(2\testdata)\right)  - \genvarlogconst(\testdata)}$ where $\customsmaller[0.9]{\genvarlogconst(\emptyset) = \log \int q_{\traindata}(z) dz = 0}$.
When inference is exact, some simple further manipulations make the two expressions equal.


Next, we specialize this result to the case of conjugate models, now assuming for simplicity that the approximate distribution lies in the conjugate family.
\begin{cor}
  \label{cor: snr monte carlo under q and conjugacy.}
  Let $p(\traindata \vert z)$ and $p(z)$ be as in \Cref{cor: snr monte carlo exp family form.}.
  Let $q_{\traindata}(z) = s(z \vert \varparamone)$ be in the conjugate family (\cref{ eq: normalization constant for conjugate prior}) with parameters $\varparamone$.
  Let $R_{K}$ be the Monte Carlo estimator for the $\PPDp$ (\cref{eq: naive mc estimator for ppdq.}.) 
  Then, $\text{SNR} (R_{K}) = \sqrt{K}/{\sqrt{\exp(\delta)^2 - 1}}$ for
  \begin{align}
  \delta 
    &=\frac{1}{2} \KL{s\pp{z\vert \varparamone + \cteststatthree{U}{}{\testdata}}}{s\pp{z\vert \varparamone}} + \frac{1}{2} \KL{s\pp{z\vert \varparamone + \cteststatthree{U}{}{\testdata}}}{s\pp{z\vert \varparamone + \cteststatthree{U}{2}{\testdata}}} \label{eq: exp fam q delta form 1} \\
     &= \frac{B\left(\varparamone\right) + B(\varparamone + \cteststatthree{U}{2}{\testdata})}{2}  - B\left(\varparamone + \cteststatthree{U}{}{\testdata}\right), \label{eq: exp fam q delta form 2}
  \end{align}
  where $B$ is as in \cref{ eq: normalization constant for conjugate prior} and
  $U(\traindata)  = \begin{bmatrix}
    T(\traindata), |\traindata|
  \end{bmatrix}
  $.
  \end{cor}
See \Cref{app: final sec: cor. approximate inference.} for a proof. This result has the same functional forms as \Cref{cor: snr monte carlo exp family form.} and differs only in the canonical parameters involved. Now, $\varparamone$ are the parameters of $q_\traindata$ and $\varparamone + U(\testdata)$ are the parameters of the posterior obtained by conditioning on $\testdata$ with $q_\traindata$ as prior.
When the inference is exact, $\varparamone = \xid$ and above expressions reduce to \Cref{cor: snr monte carlo exp family form.} expressions. Note that $\delta$ as in \cref{eq: exp fam q delta form 2} is again the looseness in Jensen's inequality: the mean of $\customsmaller[0.95]{B(\varparamone)}$ and $\customsmaller[0.95]{B(\varparamone + U(2\testdata))}$ versus $B$ applied to the mean of $\customsmaller[0.95]{\varparamone}$ and $\customsmaller[0.95]{\varparamone + U(2\testdata)}$.

\jd{I cut a lot of discussion here, feeling it didn't add much over just saying the above result specializes to the old one. Consider reverting, but I felt it was kinda a lot of words and not much "reward"}


\section{Learned Importance Sampling}
\label{sec: learned importance sampling.}

Is there anything that can be done to mitigate poor SNR? In general, when an MC estimator has high variance, a standard solution is to replace it with an importance sampling (IS) estimator \cite[Chapter~9]{mcbook}. For a valid proposal distribution $r$, the IS estimator for $\PPDp$ can be written as
\begin{align}
  R^{\mathrm{IS}}_{K} = \frac{1}{K} \sum_{k=1}^{K}\frac{p(\testdata\vert z_k) q_\traindata(z_k)}{r(z_k)}, \quad \quad \text{where } z_k \sim r(z). \label{eq: is estimator K sample}
\end{align}
The choice of the proposal distribution in crucial.
Setting $r(z) = q_\traindata(z)$ does nothing, since this reduces to the naive MC estimator.
Alternatively, one could use $\ropt \propto p(\testdata \vert z)q(z\vert \traindata)$—the IS estimator corresponding to $\ropt$ has infinite SNR and a single sample gives the exact PPD\citep{mcbook,JMLR:v21:19-102};
however, $\ropt$ is rarely tractable.

To find a tractable proposal that also provides better estimates, one could optimize an objective to learn a proposal $r_w$ with parameters $w$. 
A natural idea is to maximize the SNR of the resulting IS estimator with respect to the parameters $w$. Maximizing $\SNR{\customsmaller[0.95]{R_{K}^{\mathrm{IS}}}}$ is equivalent to minimizing the variance of $\customsmaller[0.95]{R_{K}^{\mathrm{IS}}}$,  which in turn is equivalent to minimizing the $\chi^2$-divergence between $\ropt$ and $r_w$ \citep{dieng2017variational}. 
However, recent research suggests that \emph{gradient estimators} for the $\chi^2$-divergence themselves suffer from poor SNR, making it challenging to optimize it in practice \citep{pmlr-v139-geffner21a}.

In this paper, we take an alternative approach. We consider learning a parameterized proposal $r_w$ by optimizing the importance weighted evidence lower-bound ($\IWELBO$) \citep{burda2015importance}. Let $z_m \sim r_w(z)$. Then 
\begin{align}
  \IWELBO_{M}\left[r_w(z) \parallel p(\testdata \vert z) q_{\traindata}(z)\right] 
  &\coloneqq 
  \E\left[ \log \frac{1}{M}\sum_{m=1}^{M}\frac{p(\testdata \vert z_m) q_\traindata( z_m)}{r_w(z_m)} \right]. \label{eq: iwelbo under q}
\end{align}
It is known that maximizing $\IWELBO$ in \cref{eq: iwelbo under q} is \emph{asymptotically equivalent} to minimizing the variance of $R^{\mathrm{IS}}$, or equivalently, maximizing $\SNR{R^{\mathrm{IS}}}$  \citep{maddison2017filtering,dieng2017variational,rainforth_tighter_2018,domke2018importance}. \jd{These asymptotics were already known before our paper, please (also?) cite an earlier one} More formally,
\begin{align}
  \lim_{M \to \infty } M \left(\log \PPDp - \IWELBO_{M} \right)= \left(\V [R^{\mathrm{IS}}]/2\PPDp^2\right).
\end{align}

\begin{wrapfigure}[7]{r}{0.4\textwidth}
  \vspace{-10pt}
  \centering
  \resizebox*{1.4\linewidth}{!}{
  \begin{minipage}[t][0.085\textheight]{1.5\linewidth}
    \begin{algorithmic}
    \State \texttt{LearnedIS}$(\testdata, K)$
    \vspace{2pt}
    \State \hspace{3mm} $w \leftarrow \texttt{Optimize}(\IWELBO)$
    \vspace{2pt}
    \State \hspace{3mm} $z_k \sim r_w(z) \quad \forall k \in \{1, \ldots, K\}$
    \vspace{2pt}
    \State \hspace{3mm} $R_{K}^{\mathrm{IS}} \leftarrow \frac{1}{K}\sum_{k=1}^{K}\frac{p(\testdata \vert z_k) q_\traindata(z_k)}{r_w(z_k)}$
  \end{algorithmic}
  \end{minipage}
}
\caption{ \small{Evaluating $\PPDp$ with Learned IS. \label{fig: pseudocodes}}}  
\end{wrapfigure}
So, optimizing the $\IWELBO$ in \cref{eq: iwelbo under q} can be thought of as a surrogate for optimizing the SNR of the IS estimator.
The naive gradient estimator of $\IWELBO$ also has poor SNR \citep{rainforth_tighter_2018,finke2019importance}. Fortunately, recently proposed doubly re-parameterized gradient estimator circumvents this issue \citep{tucker2018doubly,finke2019importance,pmlr-v139-bauer21a} and offers a practical option \citep{aagrawal2020}.

Overall, we propose a two step procedure to evaluate $\PPDp$.
First, learn the proposal $r_w$ by optimizing $\IWELBO$ in \cref{eq: iwelbo under q}. 
Second, use the IS estimator in \cref{eq: is estimator K sample} to evaluate the $\PPDp$. 
See \Cref{fig: pseudocodes} for a simple pseudocode of the proposed learned IS (LIS) approach.

\section{Experiments}
\label{sec:experiments}

\jd{For all examples, I favor using $K=1$ for all SNR quantities. (But $log R_K$ of course)}

\jd{In tables, need $\log R_K \rightarrow \mathbb{E} \log R_K$ and similarly for $\log R_K^{IS}$}

\jd{You cite appendix M over and over again. I think people will find this borderline enraging. Can you cite subsections? The appendix is scary long, so makes a big difference}

\jd{One issue with the examples is that you don't currently really explain anywhere why people would care about $\mathbb{E} \log R_K$, how it's commonly used to compare algorithms, etc. I don't think this is the right place though—maybe the introduction?}

We consider four settings: exponential family models, linear regression, logistic regression, and a hierarchical model. 
For first three settings, we use synthetic data sampled from the model. Such synthetic setting allow us to create different scenarios and test if the SNR problem occurs as predicted by the theory. 
For the hierarchical model, we use real-world data from MovieLens 25M dataset \citep{harper2015movielens}. The idea of this real-world settings is to simulate the use case where one uses PPD values to compare inference methods.
\jd{Cut rest of this paragraph?}

\subsection{Exponential Family Models}
\label{subsec: exponential family models.}


\begin{figure}[!h]
    
    \vspace{-12pt}
    \begin{subfigure}{0.32\linewidth}
      \includegraphics[width = \linewidth]{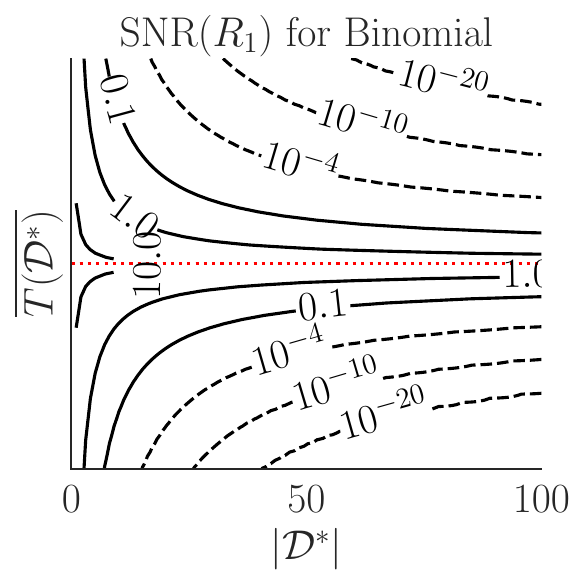}
    \end{subfigure}
    \begin{subfigure}{0.32\linewidth}
      \includegraphics[width = \linewidth]{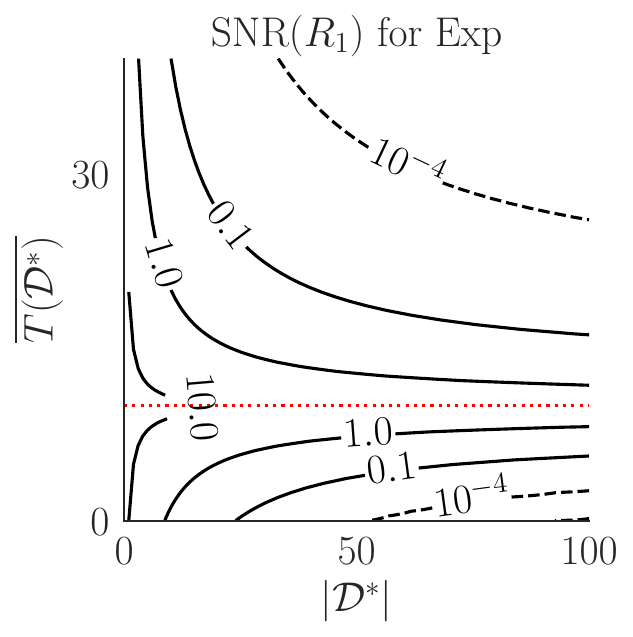}
  \end{subfigure}
  \begin{subfigure}{0.32\linewidth}
    \includegraphics[width = \linewidth]{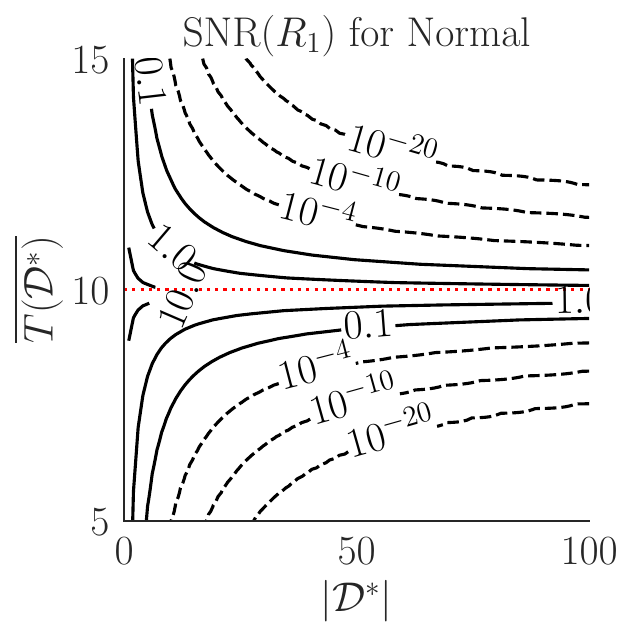}
  \end{subfigure}
  \caption{
    \small
    {\label{fig: snr for exp models.} $\SNR{R_1}$ \textbf{contours.} 
    $\overline{T}(\traindata)$ denotes the average sufficient statistics of the data points in $\traindata$.
    $\verts \traindata  = 100$ and the red dotted line indicates $\overline{T}(\traindata) = 10.$ 
    Data mismatch increases as we move away from the red dotted line, and the relative size of $\testdata$ increases as we move along the horizontal axis. Either way, SNR decreases exponentially. 
    SNR is calculated in-closed form after deriving $B$ and plugging it into $\delta$ in \cref{eq: exp family delta form 2.}. 
    \jd{Someday (for camera ready!) maybe change y axis so line doesn't go over (D)}}
  }
\end{figure}

We consider three examples of exponential family models.
First, a Normal model where $p(y \vert z)$ is a Normal distribution with known variance $\sigma^2$ and unknown mean $z \in \R$, and $p(z)$ is a Normal distribution. 
Second, an Exponential model where $p(y \vert z)$ is an Exponential distribution with unknown rate $z \in \R_{+}$, and $p(z)$ is a Gamma distribution.
Third, Binomial model where $p(y \vert z)$ is a Binomial distribution with known number of trials $n$ and unknown success probability $z \in [0,1]$, and $p(z)$ is a Beta distribution (see \Cref{tab: conjugate models.} in \Cref{app: final sec: exponential family.} for details of the models.)
\Cref{fig: snr for exp models.} shows SNR contours when inference is exact. 
For each model, we sample a dataset where the naive MC estimator suffers from low SNR and compare the performance of naive MC and learned IS estimators.

\begin{table}[!h]
  \vspace{-4pt}
  \caption{
    \label{tab: ppd estimation} 
    \small{Results $\log \PPDp$ estimation under exact inference corresponding to \Cref{tab: statistics of the data} datasets. We use $K=10^6$ for naive MC and $K=10^3$ for IS estimators. Mean and standard deviation reported over ten runs.}
  }
  \begin{center}
    \resizebox{0.75\linewidth}{!}{
      \input{./results/ppd_df.tex}
      }
  \end{center}
\end{table}
\begin{table}[!h]
  \vspace{-10pt}
  \caption{
    \label{tab: approx ppd estimation} 
    \small{Results $\log \PPDp$ estimation under approximate inference corresponding to \Cref{tab: statistics of the data} datasets. We use $K=10^6$ for naive MC and $K=10^3$ for IS estimators. Mean and standard deviation reported over ten runs.}
  }
  \begin{center}
    \resizebox{0.75\linewidth}{!}{
      \input{./results/approx_ppd_df.tex}
      }
    \end{center}
\end{table}

\jd{Need to explain what equation you're using. This reads like you're just "showing" SNR. Must be clear that you can only make this figure because of your new math! This reads like something "obvious" not a result of your core contribution. (Just 1-2 sentences needed.)}

\jd{Seems like you're now basically starting a totally separate experiment. Need to make this clear.}

\begin{wraptable}[8]{r}{0.4\textwidth}
  \caption{ \label{tab: statistics of the data} { Summary of the data sets used for results in \Cref{tab: ppd estimation,tab: approx ppd estimation}.}}
  \vspace{-15pt}
  \begin{center}
    \resizebox{\linewidth}{!}{
      \begin{tabular}[t]{@{}lrrrrr@{}}
        \toprule
        Model & $\overline{T(\traindata)}$ & $|\traindata|$ & $\overline{T(\testdata)}$ & $|\testdata|$ & $\delta$\\
        \midrule
        Normal & $10.08$ & 100 & $4.96$ & 100 & 210.85\\
        Exp & $7.00$ & 100 & $39.37$ & 100 & 11.74\\
        Binomial & $8.96$ & 100 & $41.06$ & 100& 23.32\\
        \bottomrule
      \end{tabular}
      }
    \end{center}
\end{wraptable}
\Cref{tab: statistics of the data} summarizes the statistics of the sampled datasets alongside $\delta$ values (calculated using \Cref{thm: snr monte carlo gen model form.}). 
\Cref{tab: ppd estimation} shows the results of estimating $\PPDp$ under exact inference.  
In \Cref{tab: approx ppd estimation}, we report the results from estimating $\PPDp$ under approximate inference. For the proposal and the variational families, we use full-rank Gaussian distributions. For learned IS, we use $M=16$ samples in the $\IWELBO$ and optimize for $1000$ iterations with ADAM \citep{kingma2014ADAM} and a learning rate of $0.001$. (See \Cref{app: final sec: exponential family.} for optimization details, and see \Cref{tab: expressions and computations v2} for details on computing $\log R$ and $\SNR{R}$ in \Cref{tab: ppd estimation,tab: approx ppd estimation}.)

For both exact and approximate inference, the learned IS approach outperforms naive MC. 
The empirical SNR of $R_K^{\mathrm{IS}}$ is much higher than the empirical SNR of $R_K$.
(Under exact inference, $R^{\mathrm{IS}}_1$ is deterministically equal to $\PPDp$.)
Under approximate inference, both $\log R_{K}$ and $\log R_{K}^{\mathrm{IS}}$ are lower-bounds on the true $\log \PPDp$ and learned IS lower-bound are hundreds of nats higher.

\subsection{Linear Regression}
\label{subsec: linear regression.}

\jd{need to make clear up front—you did lots of new analysis. Don't let the reader miss this!}

Consider a linear regression model where posterior is a Gaussian distribution.
We can calculate the exact SNR by plugging in the Gaussian posteriors in \Cref{thm: snr monte carlo gen model form.}. 
However, for arbitrary $\testdata$ and $\traindata$ this expression is rather complicated.
We consider a specific case where the test data $\testdata$ contains $m$ copies of the training data $\traindata$ with some mismatch and provide the following intuitive result.

\begin{thm}
  \label{thm: lin reg snr}
  Let $p(y_\traindata, z)$ be the Bayesian linear regression model.
  Let $p(y_\traindata \vert z) = \mathcal{N}(y_\traindata \vert X_\traindata z, \sigma^2 I)$ be the likelihood
  such that $y_\traindata \in \R^{\verts{\traindata}}$ is the response vector, $X_\traindata \in \R^{\verts{\traindata}\times d}$ is feature matrix, and $\sigma^2$ is the known variance. 
  Let $p(z)=\mathcal{N}(z \vert \mu_0, \Sigma_0)$ be the prior such that $z \in \R^d$. 
  Let $\traindata_\Delta$ be the mismatched copy generated by adding vector $\Delta$ to $y_\traindata$ such that $y_{\traindata_\Delta} = y_{\traindata} + \Delta$ and $X_{\traindata_\Delta} = X_{\traindata}$.
  Let $\testdata$ contain $m$ copies of $\traindata_{\Delta}$ where $m$ is a positive integer.
  Let $R_K$ be the naive Monte Carlo estimator for $\PPDp$ as in \cref{eq: naive mc estimator for ppdq.} with exact inference.
  Then, $\SNR{R_K} =\sqrt{K}/{\sqrt{\exp(\delta)^2 - 1}}$, where
  \begin{align}
    \lim_{\mathsmaller{\left(X^\top_\traindata X_\traindata\right)^{-1}\Sigma_0^{-1} \to 0}} \delta\,
      & =\, \frac{1}{2}d\log \frac{1 + m}{\sqrt{1 + 2m}} \,\, + \,\,  \frac{1}{2\sigma^2}\frac{m^2}{2m^2 + 3m + 1}\Delta^\top X_\traindata \left(X_\traindata^\top X_\traindata\right)^{-1} X_\traindata^\top \Delta
      \label{eq: lin reg snr}
  \end{align}
\end{thm}
We discuss the above result in detail in \Cref{app: final sec: linear regression.}.
The main assumption is that $\mathsmaller{\left(X_\traindata^\top X_\traindata\right)^{-1} \Sigma_0^{-1}}$ can be ignored from calculations. 
This essentially means that the feature matrix and prior parameters are such that the posterior parameters are  influenced only by data. 
This is analogous to  assumption in \Cref{prop: delta Bayesian CLT}  and RHS of \cref{eq: lin reg snr} reduces to that of \cref{eq: delta CLT} when there is no mismatch, $\Delta = 0$. Overall, $\delta$ is affected quadratically by mismatch ($\Delta$), linearly by the dimensionality of latent space ($d$), and logarithmically by the relative size of $\testdata$ ($m$).

\begin{figure}[t]
  \includegraphics[trim = 7 10 15 6, clip,  width=\textwidth]{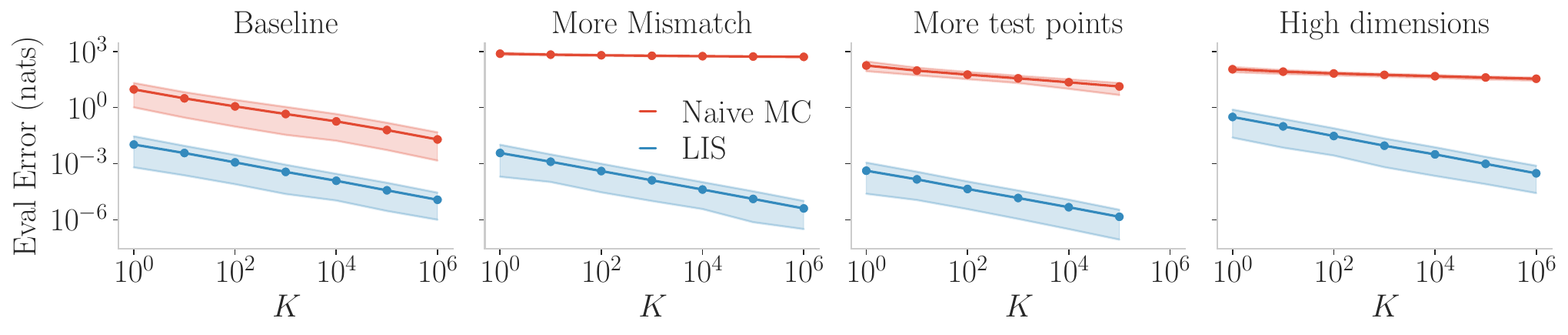}
  \caption{\small{\label{fig: lin reg estimation error} \textbf{Evaluation Error.} 
  We evaluate the error in estimating $\log \PPDp$ for the linear regression model (\cref{subsec: linear regression.}) where error is defined as $\log \PPDp - \log R_K$ and is plotted against the number of samples $K$. 
  The five and ninety-five percentiles are represented as the filled regions. 
  Across the scenarios, LIS significantly reduces the error compared to the naive estimator. See \Cref{app: final sec: linear regression.} for more details on scenarios.
  }}
\end{figure}

We use the setting as in \Cref{thm: lin reg snr} to construct synthetic datasets.
We start with a baseline where none of the three factors influencing SNR are too high. 
We then independently increase mismatch, dimensionality, and relative size to create additional scenarios (for details, see \Cref{app: final sec: linear regression.}.) 
In \Cref{fig: lin reg estimation error}, we plot the error in evaluating $\log \PPDp$ for the different scenarios using the naive MC and learned IS estimators.

For the baseline (first panel in \Cref{fig: lin reg estimation error}), the naive MC has high enough SNR and evaluation is accurate for $K = 10^6$.
The error for naive estimator increases orders of magnitude for each of the additional scenarios (see last three panels in \Cref{fig: lin reg estimation error}) compared to baseline (see red curves.)
Also, the steepness of the error slopes corresponds to the relative importance of the three factors influencing SNR of the naive estimator (see \cref{eq: lin reg snr} and red curves in \Cref{fig: lin reg estimation error}.)
We use a full-rank Gaussian to learn the proposal distribution for the learned IS estimator and optimize the $\IWELBO$ for $M=16$ with the DReG estimator and the ADAM optimizer with a learning rate of $0.001$ for $1000$ iterations. The learned IS consistently evaluates accurately across all scenarios (see blue curves in \Cref{fig: lin reg estimation error}.)
\subsection{Logistic Regression}
\label{subsec: logistic regression.}


We consider a logistic regression model with likelihood $p(y \vert z) = \mathcal{B}(y \vert \textrm{sigmoid}(z^\top x))$ and prior $p(z) = \mathcal{N}(z \vert 0, I)$ where response $y \in \{0,1\}$, latent variable $z \in \R^d$, and feature vector $x \in \R^d$.
We learn a full-rank Gaussian approximation $q_\traindata$ using variational inference. 
Unlike with linear regression, we can not calculate the SNR as in \Cref{thm: snr monte carlo under q.} because $q_{\traindata}(z \vert \testdata)$ is intractable. This prohibits an analysis similar to \Cref{thm: lin reg snr}.  
Nevertheless, we consider the case where $\testdata$ contains $m$ copies of $\traindata$ with ``mismatch''. \jd{I don't understand the next sentence at all} 

From \Cref{sec: snr problems with approximate inference.}, we know that when $q_\traindata$ is a good approximation of the true posterior, the naive MC estimator can have low SNR as mismatch, dimensionality, or the relative size of $\testdata$ increase.
We start with a baseline scenario where none of the three factors are too high and then create scenarios where each is increased one at a time.
See \Cref{app: final sec: logistic regression.} for more details.

\Cref{tab: log reg ppd estimation} reports the results from estimating $\PPDp$ using the naive MC estimator and the learned IS estimator for the different scenarios. 
We use a full-rank Gaussian proposal for the learned IS estimator and optimize the $\IWELBO$ with $M=16$ using the DReG estimator and ADAM with a learning rate of $0.001$ for $1000$ iterations.
For baseline scenario, the naive MC estimator has reasonably high SNR. Learned IS increases it much further.
For the other scenarios, naive MC estimator has low SNR, but the learned IS estimator performs very well.
\begin{table}[!h]
  \caption{\label{tab: log reg ppd estimation} 
  \small{Results for $\log \PPDp$ evaluation for logistic regression (\Cref{subsec: logistic regression.}.) We use $K = 10^6$ for naive MC and $K = 10^3$ for IS estimators. Mean and standard deviation reported over ten runs. }
  }
  \begin{center}
    \resizebox{0.8\linewidth}{!}{
      \input{./results/log_reg_ppdq.tex}
      }
    \end{center}
  \vspace{-10pt}
  \end{table}

\subsection{Hierarchical model}
\label{subsec: movielens 25m.}
MovieLens 25M \citep{harper2015movielens} is a dataset of 25 million movie ratings along with a set of features for each movie \cite{vig2012tag}. 
We randomly select 100 users after filtering those with more than 1,000 ratings. We keep one-tenth of ratings of each user as a test dataset and use remaining as a training dataset. We also PCA the movie features to ten dimensions. (See \Cref{app: final sec: hierarchical model.} for more details.)

The task is to model rating $y_{i,j} \in \{0,1\}$ of user $i$ for movie $j$ with given features $x_{i,j}$. We use a hierarchical model
$p(\theta, w, y \vert x) = \N (\theta \vert 0, I) \customsmaller[0.8]{\prod_{i=1}^{100}} \N (w_i \vert \mu(\theta), \Sigma(\theta)) \customsmaller[0.8]{\prod_{i=1}^{n_i}} \mathcal{B}(y_{i,j} \vert \mathrm{sigmoid}(w_i^\top x_{i,j})),$
where $\theta$ and $w$ together represent all the latent variables $z$; $\theta$ are the global latent variables capturing preferences over users and $w_i$ are the local latent variables capturing preferences for user $i$. $\mu$ and $\Sigma$ are functions such that if $\theta = [\theta_\mu, \theta_\Sigma]$, 
$\mu(\theta)= \theta_\mu$ and $\Sigma(\theta) = \mathrm{tril}(\theta_\Sigma)^\top \mathrm{tril}(\theta_\Sigma)$,
where $\mathrm{tril}$ takes an unconstrained vector and outputs a lower-triangular positive definite matrix. $n_i$ is the number of ratings for user $i$. $\mathcal{B}$ is the Bernoulli distribution. 

\begin{table}[!h]
  \vspace{-5pt}
  \caption{\label{tab: movielens 25m ppd estimation} 
  \small{
    Results for $\log \PPDp$ estimation for MovieLens 25M dataset (\Cref{subsec: movielens 25m.}.) Mean and standard deviation reported over ten runs.
    }
    }
    \begin{center}
      \resizebox{\linewidth}{!}{
        \input{results/movielens_results.tex}
        }
    \end{center}
    \vspace{-5pt}
\end{table}

\jd{Formerly "A common practice is to compare the inference methods using the $\PPDq$ values." this needs to be explained earlier!}

Note there is no mismatch between the training and test datasets. The relative size of test dataset $|\testdata|/|\traindata| = 0.1$ is small. The dimensionality of the latent space $d = 1065$ is high, so naive MC estimator can suffer from low SNR problem.
We consider two posterior approximations—full-rank Gaussians and normalizing flows\textemdash for learning $q_\traindata$. For flows, we use a RealNVP flow \citep{dinh2016density}  with ten affine coupling layers where the neural network in each layer has two hidden layers with $32$ units.

For the learned IS proposal, we use a normalizing flow with the same architecture as the approximate posterior. We use $M = 16$ and optimize the $\IWELBO$ for $100$ iterations with the DReG estimator and ADAM with a learning rate of $0.001$. We initialize the proposal distribution with $q_\traindata$. 

\Cref{tab: movielens 25m ppd estimation} reports the results from estimating $\log \PPDq$ using naive MC and learned IS with different values of $K$. When using the naive MC with $K = 10^6$, flow VI reports test-likelihoods more than $20$ nats higher than Gaussian VI (see the second column.)
However, the SNR of these estimates is extremely low. With learned IS flow VI is only $4$ nats higher than the Gaussian VI, and the SNR is much higher (see the fourth column.) 
So, while flow VI may be better than Gaussian VI in terms of test-likelihood, the difference is not as large as it seems when evaluated using the naive MC estimator.


\section{Discussion}
\label{sec: discussion}
\textbf{Conclusions.} We observe that the SNR of the naive PPD estimator can be extremely poor. 
We then develop intuition and theoretical understanding for the low SNR problem and demonstrate that it occurs when there is either mismatch between the training and test data, the dimensionality of the latent space is high, and/or the size of the test data is significant compared to the size of the training data.
As a secondary contribution, we propose a simple importance sampling based solution for the low SNR problem by learning a proposal distribution at test time. We show that the learned IS estimates are significantly more accurate than the naive MC.

\jd{I cut two limitations sentences. I think these either need to be elaborated or cut.}

\textbf{Limitations.}
Learned IS involves learning a proposal distribution at test time. This can be computationally expensive and may not be worth the effort when the naive MC estimator has good SNR. Future work could explore the trade-offs between the accuracy of the learned IS estimator and the computational cost of learning the proposal distribution.

\textbf{Related Works.} 
Use of annealed importance sampling (AIS) for improving posterior predictive estimates has been explored earlier \cite{wu2016quantitative,JMLR:v21:19-102,llorente2023target}.  Running MCMC procedures on approximate inference problems can be extremely slow \cite{blei2017variational}, and such methods are orthogonal to our variational approach. See \Cref{app: final sec: related works.} for detailed discussion of other related works.


\bibliography{outline}
\bibliographystyle{plainnat}
\newpage
\appendix

\section{Related Works}
\label{app: final sec: related works.}
\citet{wu2016quantitative} explored the use of Annealed Importance Sampling (AIS) \citep{neal2001annealed} for estimating the posterior predictive density in decoder based models. In particular, they used AIS for estimating the normalization constant of the unnormalized density $p(y^*_i\vert z_i)q(z_i\vert \traindata)$ for each data point $y_i$ in the test data set $\testdata$. Different from them, we focus on black-box treatment of probabilistic models \citep{ranganath14,kucukelbir2017automatic} and exploit BBVI schemes \citep{kucukelbir2017automatic,aagrawal2020} for estimating the posterior predictive densities over datasets $\testdata$. Recent theoretical advances \citep{domke2019provable,domke2020provable,domke2024provable,kim2023practical,kim2024convergence,kim2024linear} make BBVI a general purpose inference method that is reliably applicable to a wide range of problems \citep{carpenter2017stan}.  

Other research has explored learning approximate posterior distributions $q_\traindata$ to calibrate for test-time utilities \citep{stoyanov2011empirical,lopez2020decision,pmlr-v15-lacoste_julien11a,morais2022loss,kusmierczyk2019variational,knoblauch2019generalized,knoblauch2022optimization,kusmierczyk2020correcting,vadera2021post}. Such methods aim to learn a distribution $q'$ that is different from $q_\traindata$ and optimizes the expectation of some utility function under $q_\traindata$ at test-time. We focus on the problem of estimating the posterior predictive density for a given $q_\traindata$ at test-time, and do not change the given posterior; we simply focus on accurate estimation.

\citet{ruiz2016overdispersed} explored learning an importance sampling estimator for estimating the gradients for BBVI \citep{ranganath14}. They learn a proposal distribution $r$ while learning the parameters of the variational distribution $q_\traindata$, and rely on exponential families for closed-form updates. We do not focus on learning the variational distribution $q_\traindata$, and use BBVI methods for learning the proposal that can be in any suitable family of distributions \citep{rezende2015variational,papamakarios2019normalizing,webb2019improving,aagrawal2020}.

\citet{Vehtari2016} evaluate predictive accuracy using metrics that involve leave-one-out "pointwise" predictive density of the type $p(y_{i} \vert \traindata_{-i})$ over the training data $\traindata$. To estimate $p(y_i \vert \traindata_{-i}) = \int p(y_i \vert z) p(z \vert \traindata_{-i}) dz$ , the authors consider using the full posterior distribution $p(z \vert \traindata)$ as the proposal distribution. However, $p(z \vert \traindata)$ can have thinner tails than $p(z \vert \traindata_{-i})$ leading to large importance weights. To remedy this, the authors fit a Pareto distribution to the importance weights and then use statistics from the fitted distribution for final estimation. While the PSIS-LOO setting differs from our focus, one can use PSIS ideas to improve LIS estimates if $r$  is suspected of thin tails.

\citet{JMLR:v21:19-102} propose a framework for target-aware Bayesian inference (TABI) in which they decompose the posterior expectations into three components. Each of the three components is then computed as an instance of importance sampling using the Annealed Importance Sampling (AIS) or Nested Importance Sampling (NIS). One can apply the TABI framework for  $\PPDp$  estimation; however, after some simple observations, this reduces to estimating $\int p(\traindata \vert z) q_{\traindata}(z) dz $  with AIS or NIS (and is same as the approach from \citet{wu2016quantitative}.) In recent work, \citet{llorente2023target} extend the TABI framework by employing the generalized thermodynamic integration scheme (GIS) for solving the posterior expectations.
When placing these TABI approaches in context, it is crucial to note that we focus on approximate inference problems. Running MCMC procedures like AIS or thermodynamic integrations procedures like GIS is often infeasible or extremely slow on such problems (due to a large number of data points or dimensions.) Therefore, we view the MCMC procedures as an orthogonal approach to our variational approach.

\citet{pmlr-v180-reichelt22a} propose the concept of expectation programming, where a probabilistic programming system considers the target posterior expectation as a first-class citizen. They aim to build an efficient estimation pipeline when target functions are previously known. In their implementation, they currently use Annealed Importance Sampling as the choice of inference scheme. Our proposed methodology can join their suite of inference options when the target functions are more amenable to a variational formulation.

\citet{izmailov2021dangers} point out that the posteriors in Bayesian neural network can be bad at generalizing under specific dataset shifts. They uncover pathologies in the BNN posteriors that lead to poor generalization and present techniques that can possibly mitigate these. Different from them, we focus on understanding the problem of inaccurate $\PPDp$ estimation and how to improve estimation without changing the properties of the posterior.
\newpage
\section{Proof for \Cref{thm: snr monte carlo gen model form.}}
\label{app: fina sec: theorem exact inference.}
\begin{lem}
  \label{app: lem: snr of naive monte carlo estimator}
  Let $R_K$ be the Monte Carlo estimator in \cref{eq: naive mc estimator for ppdq.}. Then,
  \begin{align}
    \SNR{R_K} = \frac{\sqrt{K}}{\sqrt{\exp\left(\delta\right)^2 - 1}}, \quad \textrm{where } \delta = \frac{1}{2} \log \left(\frac{\E[R_1^2]}{\E[R_1]^2}\right)
  \end{align}
\end{lem}
\begin{proof}
  The proof follows naturally from the definition of $\SNR{R_K}$. 
  \begin{align}
    \SNR{R_K} = \sqrt{K}\SNR{R_1} &= \sqrt{K}\frac{\E[R_1]}{\sqrt{\V[R_1]}} \\
    &= \sqrt{K}\frac{\E[R_1]}{\sqrt{\E[R_1^2] - \E[R_1]^2}} \\
    &\overset{\text{(a)}}{=} \frac{\sqrt{K}}{\sqrt{ \left(\frac{\E\left[R_1^2\right]}{\E[R_1]^2} - 1\right)}}  \\
    &\overset{\text{(b)}}{=} \frac{\sqrt{K}}{\sqrt{\exp\left(2\delta\right) - 1}}  = \frac{\sqrt{K}}{\sqrt{\exp\left(\delta\right)^2 - 1}}, 
  \end{align}
  where $\text{(a)}$ follows from the fact LHS and RHS of $\overset{\text{(a)}}{=}$ are equal for $\E[R_1] > 0$ and limit is the same at $\E[R_1] = 0$; and $\text{(b)}$ follows from the definition of $\delta$.
\end{proof}

\begin{defn}[Log-normalization function]
  \label{app: def: log-normalization constant for exact inference}
  Let $\traindata$ be some dataset. Let $p(\mathcal{D} \vert z)$ be the likelihood and $p(z)$ be the prior. Then, posterior distribution $p(z \vert \traindata)= \frac{p(\traindata \vert z)p(z)}{\genposconst(\genposparam{\traindata})},$ 
  where
  \begin{align}
    \genposlogconst(\genposparam{\traindata}) \coloneqq \log \int p(\traindata \vert z) p(z) dz. \label{app: eq: normalization constant for exact inference}
  \end{align}
\end{defn}

\begin{lem}
  \label{app: lem: moments of r1}
  Let $p(\traindata \vert z)$ be the likelihood and $p(z)$ be the prior. 
  Let $\testdata$ be some test data. 
  Let $p(\traindata + \testdata \vert z) = p(\traindata \vert z) p(\testdata \vert z)$ for any $\traindata$ and $\testdata$.  
  Let $R_1$ be the Monte Carlo estimator for the $\PPDp$ under exact inference (\cref{eq: naive mc estimator for ppdq.} with $K = 1$ and $q_\traindata(z) = p(z \vert \traindata)$.) Then, 
  \begin{align}
    \E \left[R_1^c\right] = \frac{\genposconst(\genposparam{\traindata + c\testdata})}{\genposconst(\genposparam{\traindata})},
  \end{align}
  where $c$ is a non-negative integer and $V$ is as in \cref{app: def: log-normalization constant for exact inference}.
\end{lem}
\begin{proof}
  The proof is straightforward for $c = 0$.
  For $c \ge 1$, we have
  \begin{align}
    \E \left[R_1^c\right] 
    \overset{(a)}{=} \E \left[ p(\testdata \vert z)^c \right]
    &\overset{(b)}{=} \E \left[ p(c\testdata \vert z) \right]\\
    &= \int p(c\testdata \vert z) \genpostdist(z \vert \genposparam{\traindata}) dz\\
    &= \frac{\int p(c\testdata \vert z) p(\traindata \vert z) p(z) dz}{\genposconst(\genposparam{\traindata})}\\
    &\overset{(c)}{=} \frac{\int p(\traindata + c\testdata \vert z) p(z) dz}{\genposconst(\genposparam{\traindata})}\\
    &\overset{(d)}{=} \frac{\genposconst(\genposparam{\traindata + c\testdata})}{\genposconst(\genposparam{\traindata})}.
  \end{align}
  where $(a)$ follows from definition of \cref{eq: naive mc estimator for ppdq.}, $(b)$ and $(c)$ follow from the i.i.d assumption on the datasets, and $(d)$ follows from \cref{app: def: log-normalization constant for exact inference}. Note: we do not require points within a dataset to be i.i.d.
\end{proof}

\begin{lem}
  \label{app: lem: kl divergence and norm constant}
  Let $p(\traindata \vert z)$, $p(z)$, and $p(z \vert \traindata)$ be as in \cref{app: def: log-normalization constant for exact inference}. 
  Let $\traindata_a$ and $\traindata_b$ be the two multisets of data.
  Then,
  \begin{align}
    &\KL{\genpostdist(z \vert \genposparam{\traindata_a})}{\genpostdist(z \vert \genposparam{\traindata_b})} \\
    & = \E \left[\log \frac{p(\traindata_a \vert z)}{p(\traindata_b \vert z)}\right]  - \genposlogconst(\traindata_a) + \genposlogconst(\traindata_b)\\
  \end{align}
\end{lem}
\begin{proof}
  \begin{align}
    &\KL{\genpostdist(z \vert \genposparam{\traindata_a})}{\genpostdist(z \vert \genposparam{\traindata_b})} \\
    & = \E \left[\log \frac{\genpostdist(z \vert \genposparam{\traindata_a})}{\genpostdist(z \vert \genposparam{\traindata_b})}\right]\\
    & = \E \left[\log \frac{\frac{p(\traindata_a \vert z)p(z)}{\exp(\genposlogconst(\traindata_a))}}{\frac{p(\traindata_b)p(z)}{\exp \genposlogconst(\traindata_b)}}\right]\\
    & = \E \left[\log \frac{p(\traindata_a \vert z)}{p(\traindata_b \vert z)}\right]  - \genposlogconst(\traindata_a) + \genposlogconst(\traindata_b)\\
  \end{align}
\end{proof}

\begin{lem}
  \label{app: lem: kl divergence and norm constant pairs}
  Let $p(\traindata \vert z)$, $p(z)$, and $p(z \vert \traindata)$ be as in \cref{app: def: log-normalization constant for exact inference}. 
  Let $\traindata_1$, $\traindata_2$, and $\traindata_3$ be the three multisets of data.
  Then,
  \begin{align}
    &\frac{1}{2}\KL{\genpostdist(z \vert \genposparam{\traindata_3})}{\genpostdist(z \vert \genposparam{\traindata_1})} 
    + \frac{1}{2}\KL{\genpostdist(z \vert \genposparam{\traindata_3})}{\genpostdist(z \vert \genposparam{\traindata_2})} \\
    & = \E \left[\log p(\traindata_3 \vert z) - \frac{1}{2}\log p(\traindata_1 \vert z) - \frac{1}{2}\log p(\traindata_2 \vert z)\right]  \\
    & + \frac{V(\traindata_1)  + V(\traindata_2)}{2} - V(\traindata_3).
  \end{align}
\end{lem}
\begin{proof}
  Applying \cref{app: lem: kl divergence and norm constant} to $\traindata_3$ and $\traindata_1$ gives 
  \begin{align}
    &\KL{\genpostdist(z \vert \genposparam{\traindata_3})}{\genpostdist(z \vert \genposparam{\traindata_1})} \\
    & = \E \left[\log p(\traindata_3 \vert z) - \log p(\traindata_1 \vert z)\right]  - \genposlogconst(\traindata_3) + \genposlogconst(\traindata_1)\\
  \end{align}
  and applying it to $\traindata_3$ and $\traindata_2$ gives
  \begin{align}
    &\KL{\genpostdist(z \vert \genposparam{\traindata_3})}{\genpostdist(z \vert \genposparam{\traindata_2})} \\
    & = \E \left[\log p(\traindata_3 \vert z) - \log p(\traindata_2 \vert z)\right]  - \genposlogconst(\traindata_3) + \genposlogconst(\traindata_2).\\
  \end{align}
  Now, multiplying the above two equations by $\frac{1}{2}$ and adding them gives
  \begin{align}
    &\frac{1}{2}\KL{\genpostdist(z \vert \genposparam{\traindata_3})}{\genpostdist(z \vert \genposparam{\traindata_1})} 
    + \frac{1}{2}\KL{\genpostdist(z \vert \genposparam{\traindata_3})}{\genpostdist(z \vert \genposparam{\traindata_2})} \\
    & = \frac{1}{2}\E \left[\log p(\traindata_3 \vert z) - \log p(\traindata_1 \vert z)\right]  - \frac{1}{2}\genposlogconst(\traindata_3) + \frac{1}{2}\genposlogconst(\traindata_1)\\
    & + \frac{1}{2}\E \left[\log p(\traindata_3 \vert z) - \log p(\traindata_2 \vert z)\right]  - \frac{1}{2}\genposlogconst(\traindata_3) + \frac{1}{2}\genposlogconst(\traindata_2)\\
    & = \E \left[\log p(\traindata_3 \vert z) - \frac{1}{2}\log p(\traindata_1 \vert z) - \frac{1}{2}\log p(\traindata_2 \vert z)\right]  \\
    & + \frac{\genposlogconst(\traindata_1)  + \genposlogconst(\traindata_2)}{2} - \genposlogconst(\traindata_3).
  \end{align}
\end{proof}

\begin{cor}
  \label{app: cor: gen thm kl divergence}
  Let $p(\traindata \vert z)$, $p(z)$, and $p(z \vert \traindata)$ be as in \cref{app: def: log-normalization constant for exact inference}. Let $\traindata_1 = c_a\traindata$, $\traindata_2 = c_a\traindata + 2c_b \testdata$, and $\traindata_3 = c_a\traindata + c_b\testdata$ be the three multisets of data where $c_a$ and $c_b$ are non-negative integers. Then,
  \begin{align}
    &\frac{1}{2}\KL{\genpostdist(z \vert \genposparam{\traindata_3})}{\genpostdist(z \vert \genposparam{\traindata_1})} 
    + \frac{1}{2}\KL{\genpostdist(z \vert \genposparam{\traindata_3})}{\genpostdist(z \vert \genposparam{\traindata_2})} \\
    & =\frac{\genposlogconst(\traindata_1)  + \genposlogconst(\traindata_2)}{2} - \genposlogconst(\traindata_3).
  \end{align}

\end{cor}

\begin{thm}[Repeated for convenience]
  Let $p(\traindata \vert z)$ be the likelihood and $p(z)$ be the prior. Let $\testdata$ be some test data. 
  Let $p(\traindata + \testdata \vert z) = p(\traindata \vert z) p(\testdata \vert z)$ for any $\traindata$ and $\testdata$. 
  Let $R_K$ (as in \cref{eq: naive mc estimator for ppdq.}) be the Monte Carlo estimator for the $\PPDp$ under exact inference. Then, the signal-to-noise ratio of $R_K$ is given by
  $\SNR{R_K} = \sqrt{K}/\sqrt{\exp(\delta)^2 - 1}$ where
  \begin{align}
    \delta 
    & = \frac{1}{2} \KL{\genpostdist(z \vert \genposparam{\traindata + \testdata})}{\genpostdist(z \vert \genposparam{\traindata})} + \KL{\genpostdist(z \vert \genposparam{\traindata + \testdata})}{\genpostdist(z \vert \genposparam{\traindata + 2\testdata})} \label{app: eq: gen them delta in kl}  \\ 
    & = \frac{\genposlogconst(\genposparam{\traindata }) + \genposlogconst(\genposparam{\traindata + 2\testdata})}{2}  - \genposlogconst(\genposparam{\traindata + \testdata})  \label{app: eq: gen them delta in constant}
  \end{align}
  where $\genposlogconst$ is as in \cref{app: def: log-normalization constant for exact inference}.
\end{thm}
\begin{proof}[Proof sketch] 
  A simple calculation gives $\text{SNR}(R_1)=\sqrt{K}/\sqrt{\exp(\delta)^2-1}$ where $\delta = \frac{1}{2} \log \E[R_1^2] - \log \E[R_1]^2$ for any single-sample unbiased estimator $R_1$ (see \cref{app: lem: snr of naive monte carlo estimator}). 
  From the i.i.d. assumption over the datasets and the likelihood, we get $\E[R_{1}^c] = \genposconst(\genposparam{\traindata + c\testdata})/\genposconst(\genposparam{\traindata})$ for all non-negative integers $c$ and $V = \log \int p(\traindata \vert z) p (z) dz $ is as in \cref{app: def: log-normalization constant for exact inference}  (see \cref{app: lem: moments of r1}.) 
  Using this with $c=1$ and $c=2$ and simplifying gives \cref{app: eq: gen them delta in constant}.  
  Then, we identify a relationship between KL-divergence between two posteriors in terms of the likelihood rations and the log-normalization constants (see \cref{app: lem: kl divergence and norm constant}.) 
  Applying this to each of the KL divergences in \cref{app: eq: gen them delta in kl} and averaging gives the same expression as in \cref{app: eq: gen them delta in constant}.
\end{proof}

  \begin{proof}
      \begin{align}
        \delta 
        &\overset{(a)}{=} \frac{1}{2} \log \frac{\E[R_1^2]}{\E[R_1]^2} \\
        &\overset{}{=} \frac{1}{2} \log \E \left[R_1^2\right] - \log \E \left[R_1\right]\\
        &\overset{(b)}{=} \frac{1}{2} \log \frac{\genposconst(\genposparam{\traindata + 2\testdata})}{\genposconst(\genposparam{\traindata})} - \log \frac{\genposconst(\genposparam{\traindata + \testdata})}{\genposconst(\genposparam{\traindata })}\\
        &\overset{(c)}{=} \frac{\genposlogconst(\genposparam{\traindata + 2\testdata})  + \genposlogconst(\genposparam{\traindata })}{2}  - \genposlogconst(\genposparam{\traindata + \testdata}) 
      \end{align}
    (a) follows from \cref{app: lem: snr of naive monte carlo estimator}, (b) follows from \cref{app: lem: moments of r1}, and (c) follows from some simple algebraic manipulations. 
    Now, for the KL-divergence result, if we take the expression in \cref{app: cor: gen thm kl divergence}, and  plug $\traindata_1 = \traindata$ and $\traindata_2 = \traindata + 2 \testdata$ and $\traindata_3 = \traindata + \testdata$, then we get the same expression as \cref{app: eq: gen them delta in constant}.
\end{proof}

\newpage

\section{Proof for \Cref{prop: delta Bayesian CLT}}
\label{app: final sec: prop.}
\begin{lem}
  \label{app: lem: KL for gaussians}
  Let $\N\pp{\mu_{0},\Sigma_{0}}$ and $\N\pp{\mu_{1},\Sigma_{1}}$ be two Gaussian distributions of dimensionality $d$ with $\Sigma_{0},\Sigma_{1}\succ0$. Then,
  \begin{align}
    \KL{\N\pp{\mu_{0},\Sigma_{0}}}{\N\pp{\mu_{1},\Sigma_{1}}}
    &=\tr\pars{\frac{1}{2}\Sigma_{1}^{-1}\Sigma_{0}}-\frac{1}{2}d\nonumber\\
    &\phantom{blah}+\frac{1}{2}\pp{\mu_{1}-\mu_{0}}^{\top}\Sigma_{1}^{-1}\pp{\mu_{1}-\mu_{0}}\nonumber\\
    &\phantom{blah}+\frac{1}{2}\ln\verts{\det\Sigma_{1}}-\frac{1}{2}\ln\verts{\det\Sigma_{0}}.
   \end{align}
  \end{lem}

\begin{cor}
  \label{app: cor: KL for two gaussians}
  Let $\N\pp{\mu_{0},\Sigma_{0}}$, $\N\pp{\mu_{1},\Sigma_{1}}$, and $\N\pp{\mu_{2},\Sigma_{2}}$ be three Gaussian distributions of dimensionality $d$ with $\Sigma_{0},\Sigma_{1},$ and $\Sigma_{2}\succ0$. Then,
  \begin{eqnarray*}
    &  & \KL{\N\pp{\mu_{0},\Sigma_{0}}}{\N\pp{\mu_{1},\Sigma_{1}}}\\
    &  &+\KL{\N\pp{\mu_{0},\Sigma_{0}}}{\N\pp{\mu_{2},\Sigma_{2}}}\\
    &  &= \tr\pars{\pars{\frac{1}{2}\Sigma_{1}^{-1}+\frac{1}{2}\Sigma_{2}^{-1}}\Sigma_{0}}-d\\
    &  &+ \frac{1}{2}\pp{\mu_{1}-\mu_{0}}^{\top}\Sigma_{1}^{-1}\pp{\mu_{1}-\mu_{0}}+\frac{1}{2}\pp{\mu_{2}-\mu_{0}}^{\top}\Sigma_{2}^{-1}\pp{\mu_{2}-\mu_{0}}\\
    &  &+ \frac{1}{2}\ln\verts{\det\Sigma_{1}}+\frac{1}{2}\ln\verts{\det\Sigma_{2}}-\ln\verts{\det\Sigma_{0}}
   \end{eqnarray*}
\end{cor}

\begin{proposition}[Repeated]
    \label{app: prop: delta Bayesian CLT}
    Suppose $\testdata$ and $\traindata$ are large enough that posteriors in \cref{eq: new delta form 1.} are well-approximated via the Bayesian CLT as Gaussians centered at their maximum-likelihood estimates (MLEs). Also suppose that $\traindata$, $\traindata + \testdata$, and $\traindata + 2\testdata$ are similar enough that the MLE and Hessian of the \emph{average} log-likelihood is the same for all three. If $d$ is the number of dimensions of $z$, then
    \begin{align}
    \delta \approx \frac{d}{2} \log \frac{1 + \verts{\testdata}/\verts{\traindata}}{\sqrt{1 + 2 \verts{\testdata} / \verts{\traindata}}}.\label{app: eq: delta CLT}
    \end{align}
  \end{proposition}
\begin{proof}
For any dataset $\traindata$, let $\hat{z}_{\traindata}$
be the maximum likelihood estimate and $-S_{\traindata}^{-1}$ be
the Hessian evaluated at the maximum likelihood estimate $\nabla_{z}^{2}\log p\pp{\traindata\vert\hat{z}_{\traindata}}$, such that,
\begin{align}
  \hat{z}_{\traindata}  =  \argmax_{z}\log p\pp{\traindata\vert z}, \quad \quad \text{ and } \quad \quad  S_{\traindata}^{-1}  =  -\nabla_{z}^{2}\log p\pp{\traindata\vert\hat{z}_{\traindata}}.
\end{align}
When $\testdata$ and $\traindata$ are large, using Bayesian central limit theorem, we can approximate all three distributions in \cref{eq: new delta form 1.} as 
\begin{align}
p\pp{z\vert \traindata} & \approx  \N\pars{z\vert\hat{z}_{\traindata},S_{\traindata}},\\
p\pp{z\vert \traindata+\testdata} & \approx  \N\pars{z\vert\hat{z}_{\traindata+\testdata},S_{\traindata+\testdata}}, \quad \text{ and }\\
p\pp{z\vert \traindata+2\testdata} & \approx  \N\pars{z\vert\hat{z}_{\traindata+2\testdata},S_{\traindata+2\testdata}}.
\end{align}

With the above approximations, we can use \Cref{app: cor: KL for two gaussians} to simplify the sum of KL-divergences appearing in \cref{eq: new delta form 2.} as follows.
\begin{align}
   & \KL{\N\pars{\hat{z}_{\traindata+\testdata},S_{\traindata+\testdata}}}{\N\pars{\hat{z}_{\traindata},S_{\traindata}}}\nonumber\\
   & +\KL{\N\pars{\hat{z}_{\traindata+\testdata},S_{\traindata+\testdata}}}{\N\pars{\hat{z}_{\traindata+2\testdata},S_{\traindata+2\testdata}}}\nonumber\\
 = & \tr\pars{\pars{\frac{1}{2}S_{\traindata}^{-1}+\frac{1}{2}S_{\traindata+2\testdata}^{-1}}S_{\traindata+\testdata}}-d\nonumber\\
 + & \frac{1}{2}\pp{\hat{z}_{\traindata}-\hat{z}_{\traindata+\testdata}}^{\top}S_{\traindata}^{-1}\pp{\hat{z}_{\traindata}-\hat{z}_{\traindata+\testdata}}+\frac{1}{2}\pp{\hat{z}_{\traindata+2\testdata}-\hat{z}_{\traindata+\testdata}}^{\top}S_{\traindata+2\testdata}^{-1}\pp{\hat{z}_{\traindata+2\testdata}-\hat{z}_{\traindata+\testdata}}\nonumber\\
 + & \frac{1}{2}\ln\verts{\det S_{\traindata}}+\frac{1}{2}\ln\verts{\det S_{\traindata+2\testdata}}-\ln\verts{\det S_{\traindata+\testdata}}. \label{app: eq: KL for two gaussians substituted}
\end{align}

From the assumption in the proposition, we have 
\begin{align}
  \hat{z}_{\traindata} = \hat{z}_{\traindata+\testdata} = \hat{z}_{\traindata+2\testdata} \label{app: eq: similar means}.
\end{align}
Also, the MLE is the same and we assume data sets to be similar, we expect the scaled Hessian to be the same, such that,
\begin{align}
  \frac{1}{\verts \traindata}S_{\traindata}^{-1} \approx \frac{1}{\verts{\traindata+\testdata}}S_{\traindata+\testdata}^{-1} \approx \frac{1}{\verts{\traindata+2\testdata}}S_{\traindata+2\testdata}^{-1} \label{app: eq: similar covariances}.
\end{align}

Substituting from \cref{app: eq: similar means,app: eq: similar covariances} into \cref{app: eq: KL for two gaussians substituted}, and simplifying as in \Cref{app: sec: note for the simplification}, we get
\begin{align}
&\KL{\N\pars{\hat{z}_{\traindata+\testdata},S_{\traindata+\testdata}}}{\N\pars{\hat{z}_{\traindata},S_{\traindata}}}\nonumber\\
&+\KL{\N\pars{\hat{z}_{\traindata+\testdata},S_{\traindata+\testdata}}}{\N\pars{\hat{z}_{\traindata+2\testdata},S_{\traindata+2\testdata}}}\nonumber\\
&\approx d\log\frac{\verts{\traindata+\testdata}}{\sqrt{\verts \traindata\verts{\traindata+2\testdata}}}. \label{app: eq: KL for two gaussians substituted 2}
\end{align}

Finally, plugging the KL-divergences from \cref{app: eq: KL for two gaussians substituted 2} into the definition of $\delta$ in \cref{eq: new delta form 1.}, we get the result
\begin{align}
  \delta
  &\approx \frac{1}{2}d\log\frac{\verts{\traindata+\testdata}}{\sqrt{\verts \traindata\verts{\traindata+2\testdata}}}=\frac{1}{2}d\log\frac{1+\verts{\testdata}/\verts \traindata}{\sqrt{1+2\verts{\testdata}/\verts \traindata}},
\end{align}
  
where the middle term shows that $\delta$ is positive\textemdash the quantity
inside the logarithm is larger than one since $\verts{\traindata+\testdata}$ is the
arithmetic mean of $\verts \traindata$ and $\verts{\traindata+2\testdata}$ which is always
larger than the geometric mean $\sqrt{\verts \traindata\verts{\traindata+2\testdata}}$.
The right term clarifies that only the dimensionality and ratio of
$\verts \traindata$ and $\verts{\testdata}$ that matters.

\end{proof}
\subsection{Note for the simplification from \cref{app: eq: KL for two gaussians substituted} to \cref{app: eq: KL for two gaussians substituted 2}}
\label{app: sec: note for the simplification}
When  the datasets $\testdata$ and $\traindata$ have the matching mean statistics, we have the relations in \cref{app: eq: similar means,app: eq: similar covariances}. Under \cref{app: eq: similar means}, the quadratic terms in \cref{app: eq: KL for two gaussians substituted} are zero. We can simplify the term involving trace as follows:
\begin{eqnarray*}
& &\tr\pars{\pars{\frac{1}{2}S_{\traindata}^{-1}+\frac{1}{2}S_{\traindata+2\testdata}^{-1}}S_{\traindata+\testdata}} \\
&  &= \frac{1}{2}\tr\pars{S_{\traindata}^{-1}S_{\traindata+\testdata}}+\frac{1}{2}\tr\pars{S_{\traindata+2\testdata}^{-1}S_{\traindata+\testdata}}\\
 & &\overset{\text{(a)}}{\approx} \frac{1}{2}\tr\pars{S_{\traindata}^{-1}\pars{\frac{\verts{\traindata+\testdata}}{\verts \traindata}S_{\traindata}^{-1}}^{-1}} +\frac{1}{2}\tr\pars{\pars{\frac{\verts{\traindata+2\testdata}}{\verts \traindata}S_{\traindata+2\testdata}^{-1}}\pars{\frac{\verts{\traindata+\testdata}}{\verts \traindata}S_{\traindata}^{-1}}^{-1}}\\
 &  &= \frac{1}{2}\frac{\verts \traindata}{\verts{\traindata+\testdata}}\tr\pars{S_{\traindata}^{-1}S_{\traindata}}+\frac{1}{2}\frac{\verts{\traindata+2\testdata}}{\verts \traindata}\frac{\verts \traindata}{\verts{\traindata+\testdata}}\tr\pars{S_{\traindata}^{-1}S_{\traindata}}\\
 &  &= \frac{1}{2}\frac{\verts \traindata}{\verts{\traindata+\testdata}}d+\frac{1}{2}\frac{\verts{\traindata+2\testdata}}{\verts{\traindata+\testdata}}d\\
 & &\overset{\text{(b)}}{=} \frac{\frac{1}{2}\verts \traindata+\frac{1}{2}\verts{\traindata+2\testdata}}{\verts{\traindata+\testdata}}d\\
 &  &= \frac{\verts{\traindata+\testdata}}{\verts{\traindata+\testdata}}d\\
 &  &= d,
\end{eqnarray*}
where (a) follows from the fact that relation in \cref{app: eq: similar covariances}; and (b) follows from the multiset notation \citep{costa2021introduction}. 

Therefore, the first and the second term ($d$ and $-d$) in \cref{app: eq: KL for two gaussians substituted} cancel out and the only remaining terms are the ones involving the logarithms of the determinants of the covariance matrices. These remaining terms can be simplified as follows:
\begin{eqnarray*}
 &  & \frac{1}{2}\ln\det\pars{S_{\traindata}}+\frac{1}{2}\ln\det\pars{S_{\traindata+2\testdata}}-\ln\det\pars{S_{\traindata+\testdata}}\\
 & \overset{\text{(c)}}{\approx} & \frac{1}{2}\ln\det\pars{S_{\traindata}}+\frac{1}{2}\ln\det\pars{\frac{\verts \traindata}{\verts{\traindata+2\testdata}}S_{\traindata}}-\ln\det\pars{\frac{\verts \traindata}{\verts{\traindata+\testdata}}S_{\traindata}}\\
 & \overset{\text{(d)}}{=} & \frac{1}{2}\ln\det\pars{S_{\traindata}}+\frac{1}{2}\ln\det\pars{S_{\traindata}}+\frac{d}{2}\log\pars{\frac{\verts \traindata}{\verts{\traindata+2\testdata}}}-\ln\det\pars{S_{\traindata}}-d\log\pars{\frac{\verts \traindata}{\verts{\traindata+\testdata}}}\\
 & = & \frac{d}{2}\log\pars{\frac{\verts \traindata}{\verts{\traindata+2\testdata}}}-d\log\pars{\frac{\verts \traindata}{\verts{\traindata+\testdata}}}\\
 & \overset{\text{(f)}}{=} & d\pars{\log\verts{\traindata+\testdata}-\frac{1}{2}\log\verts \traindata-\frac{1}{2}\log\verts{\traindata+2\testdata}}\\
 & = & d\log\frac{\verts{\traindata+\testdata}}{\sqrt{\verts \traindata\verts{\traindata+2\testdata}}},\\
\end{eqnarray*}
where (f) follows from \cref{app: eq: similar covariances}; (d) follows from $\log\det\pars{aX}=d\log a+\log\det X$ for any non-negative scalar $a$; this gives the final result in \cref{app: eq: KL for two gaussians substituted 2}; and (c) follows from simple algebraic manipulations.

\newpage

\section{Proof for \Cref{cor: snr monte carlo exp family form.}}
\label{app: final sec: cor. exact inference}

\begin{lem}
  \label{app: lem: moments of the estimator}
  Let the likelihood $p(y\vert z)$ be in exponential family (\cref{eq: conjugate likelihood}) and prior $p(z) = s(z \vert \xi_0)$ be in the corresponding conjugate family (\cref{ eq: normalization constant for conjugate prior}). Let $\traindata$ be a multiset of training data, $\testdata$ a multiset of test data, and let $R_1$ be the Monte Carlo estimator for the PPD with exact inference (\cref{eq: naive mc estimator for ppdq.} with $K = 1$). Let $\displaystyle h(\testdata) = \prod_{y \in \testdata} h(y)$. Then, 
  \begin{align}
    \E[R_1]^c &= h(\testdata)^c \frac{\exp B(\traindata + c \testdata)}{\exp B(\traindata)}, 
  \end{align}
  where $c$ is a non-negative integer and $B$ is as in \cref{ eq: normalization constant for conjugate prior}.
\end{lem}
\begin{proof}
  Starting from the definition of $R_1$ we have,
  \begin{align}
    \E[R_1^c] 
    = \E\left[\left(p(\testdata\vert z)\right)^c\right]
    &
    =\E\left[\left(\prod_{y \in \testdata} p(y \vert z)\right)^c\right]\\
    &=\E\left[\left(\prod_{y \in \testdata} h(y) \exp\left(T(y)^\top \phi(z) - A(z)\right)\right)^c\right]\\
    &\overset{\textrm{(a)}}{=} \E\left[\left(h(\testdata) \exp\left(T(\testdata)^\top \phi(z)  - |\testdata|A(z)\right)\right)^c\right],\\
  \end{align}
  where $\text{(a)}$ follows from $T(\testdata) = \sum_{y \in \testdata} T(y)$ and $h(\testdata) = \prod_{y\in \testdata} h(y)$. Doing some basic manipulations, we get
  \begin{align}
    &\E\left[\left(h(\testdata) \exp\left(T(\testdata)^\top \phi(z)  - |\testdata|A(z)\right)\right)^c\right]\\
    &= h(\testdata)^c \E\left[\exp\left(cT(\testdata)^\top \phi(z)  - c|\testdata|A(z)\right)\right]\\
    &\overset{\textrm{(b)}}{=} h(\testdata)^c \int \exp\left(cT(\testdata)^\top \phi(z)  - c|\testdata|A(z)\right) s(z \vert \xi_\traindata) dz\\
    &\overset{\textrm{(c)}}{=} h(\testdata)^c \frac{\int \exp\left(cT(\testdata)^\top \phi(z)  - c|\testdata|A(z) \right) \exp \left(T(\traindata)^\top \phi(z) - |\traindata|A(z)\right) dz}{\exp(B(\xid))}\\
    &\overset{\textrm{(d)}}{=} h(\testdata)^c \frac{\int \exp\left(T(\traindata + c\testdata)^\top \phi(z)  - (|\traindata + c\testdata|)A(z)\right) dz}{\exp(B(\xid))}\\
    &\overset{\textrm{(e)}}{=} h(\testdata)^c \frac{\exp(B(\xi_{\traindata + c \testdata}))}{\exp(B(\xid))}\\
  \end{align}
  where (b) and (c) follow from the definition of $s(z \vert \xid)$ (\cref{ eq: normalization constant for conjugate prior}) and the fact that the expectation is under the posterior; $\text{(d)}$ follows from the the multiset notation \cite{costa2021introduction}; $\text{(e)}$ follows from the definition of $B$ in \cref{ eq: normalization constant for conjugate prior}.
\end{proof}

\begin{lem}
  \label{app: lem: kl divergence of two exponential family distributions}
  In a canonical exponential family $p\pp{x\vert \eta}=h\pp x\exp\pars{T\pp x^{\top}\eta-A\pp \eta},$
  the looseness of Jensen's equality applied to the log-partition function $A$ at points $v,w,$ and $u = \frac{v + w}{2}$
  is
  \[
  \frac{1}{2}\pars{A\pp v+A\pp w}-A\pp u=\frac{1}{2}\KL{p\pp{x\vert u}}{p\pp{x\vert v}}+\frac{1}{2}\KL{p\pp{x\vert u}}{p\pp{x\vert w}}.
  \]
  \end{lem}
  
  \begin{proof}
  The KL-divergence between two canonical exponential family distributions with parameters $v$ and $w$ is given by
  \begin{eqnarray}
  \KL{p\pp{x\vert w}}{p\pp{x\vert v}} = \E_{p\pp{x\vert w}}\log\frac{p\pp{x\vert w}}{p\pp{x\vert v}}
   & = & \E_{p\pp{x\vert w}}\pars{T\pp x^{\top}w-T\pp x^{\top}v-A\pp w+A\pp v}\\
   & = & \pars{w-v}^{\top}\E_{p\pp{x\vert w}}\bracs{T\pp x}-A\pp w+A\pp v\\
   & \overset{\text{(a)}}{=} & \pars{w-v}^{\top}\nabla A\pp w-A\pp w+A\pp v, \label{app: eq: kl divergence in canonical exponential family 1}
  \end{eqnarray}
  where $\text{(a)}$ follows from the definition of the gradient of $A$. 
  
  Now, rearranging terms in \cref{app: eq: kl divergence in canonical exponential family 1} gives an expression for log-partition function $A$ at any point $w$ in terms of the log-partition function $A$ at any other point $v$ and the KL-divergence between the two distributions:
  \begin{eqnarray}
    A\pp w & = & A\pp v+\pars{w-v}^{\top}\nabla A\pp w-\KL{p\pp{x\vert w}}{p\pp{x\vert v}} \label{app: eq: kl divergence in canonical exponential family}
  \end{eqnarray}
  Replacing $w$ with $u$ in \cref{app: eq: kl divergence in canonical exponential family}, gives 
  \begin{eqnarray}
  A\pp u & = & A\pp v+\pars{u-v}^{\top}\nabla A\pp u-\KL{p\pp{x\vert u}}{p\pp{x\vert v}}, \label{app: eq: kl divergence in canonical exponential family 2}
  \end{eqnarray}
  and replacing $w$ with $u$ and $v$ with $w$ in \cref{app: eq: kl divergence in canonical exponential family} gives
  \begin{eqnarray}
  A\pp u & = & A\pp w+\pars{u-w}^{\top}\nabla A\pp u-\KL{p\pp{x\vert u}}{p\pp{x\vert w}}. \label{app: eq: kl divergence in canonical exponential family 3}
  \end{eqnarray}
  On averaging \cref{app: eq: kl divergence in canonical exponential family 2} and \cref{app: eq: kl divergence in canonical exponential family 3} the $\nabla A(u)$ terms cancel out and we get
  \begin{eqnarray}
  A\pp u & = & 
  \frac{1}{2}\pars{A\pp v+A\pp w}\nonumber\\
  & & -\frac{1}{2}\KL{p\pp{x\vert u}}{p\pp{x\vert v}}-\frac{1}{2}\KL{p\pp{x\vert u}}{p\pp{x\vert w}}
\end{eqnarray}
  Finally, rearranging the terms, proves the result:
  \begin{eqnarray}
    \frac{1}{2}\pars{A\pp v+A\pp w}-A\pp u & = & \frac{1}{2}\pars{\KL{p\pp{x\vert u}}{p\pp{x\vert v}}+\KL{p\pp{x\vert u}}{p\pp{x\vert w}}}.
  \end{eqnarray}
  \end{proof}

\begin{thm}[Repeated]\label{app: thm: snr monte carlo}
  Take a model with a likelihood $p(y \vert z)$ in an exponential family (\cref{eq: conjugate likelihood}) and a prior $p(z)=s(z \vert \xi_0)$ in the corresponding conjugate family (\cref{ eq: normalization constant for conjugate prior}). 
  Let $\testdata$ be some test data. 
  Let $R_K$ be the Monte Carlo estimator for the PPD under exact inference (\cref{eq: naive mc estimator for ppdq.}.) 
  Then, the signal to noise ratio is $\text{SNR} (R_K) = \sqrt{K}/\sqrt{\exp\left(\delta\right)^2 - 1}$ for
  \begin{align}
  \delta &= \frac{1}{2}\KL{s\pp{z\vert \traindata+\testdata}}{s\pp{z\vert \traindata}}+\frac{1}{2}\KL{s\pp{z\vert \traindata+\testdata}}{s\pp{z\vert \traindata+2\testdata}} \label{app: eq: delta KL} \\
     &= \frac{B\left(\xi_\traindata\right) + B(\xi_{\traindata + 2\testdata})}{2}  - B\left(\xi_{\traindata + \testdata}\right),\label{app: eq: delta} 
  \end{align}
  where for any dataset $\traindata$, $\xi_\traindata$ are the parameters that make the conjugate family $s(z \vert \xi_\traindata)$ equal to the posterior density $p(z \vert \traindata)$ (\cref{eq: xiD}), and $B$ is as in \cref{ eq: normalization constant for conjugate prior}.
  \end{thm}
  \begin{proof}
    From \Cref{app: lem: snr of naive monte carlo estimator} we get $\SNR{R_K} = \frac{\sqrt{K}}{\sqrt{\exp(\delta)^2 - 1}}$ for $\delta = \frac{1}{2} \log(\E[R_1^2]/\E[R_1]^2)$. Using \Cref{app: lem: moments of the estimator}, for $c=1$ and $c=2$, we can simplify $\delta$ as 
    \begin{align}
      \delta 
      = \frac{1}{2} \log \frac{\E[R_1^2]}{\E[R_1]^2}
      &= \frac{1}{2} \log \E\left[R_1^2\right] - \log \E\left[R_1\right]\\
      &\overset{\textrm{(a)}}{=} \frac{1}{2} \log h(\testdata)^2 \frac{\exp B(\traindata + 2\testdata)}{\exp B(\traindata)} - \log h(\testdata) \frac{\exp B(\traindata + \testdata)}{\exp B(\traindata)}\\
      &\overset{\textrm{(b)}}{=} \frac{1}{2} \log \frac{\exp B(\traindata + 2\testdata)}{\exp B(\traindata)} - \log \frac{\exp B(\traindata + \testdata)}{\exp B(\traindata)}\\
      &\overset{\textrm{(c)}}{=} \frac{B(\xi_{\traindata + 2\testdata}) + B(\xi_\traindata)}{2} - B(\xi_{\traindata + \testdata}),
    \end{align}
    where (a) follows from \Cref{app: lem: moments of the estimator} for $c =1$ and $c = 2$, (b) follows from cancellations of $\log h(\testdata)$, and (c) follows from simple algebra.
    
    Now, observe $B$ in \cref{ eq: normalization constant for conjugate prior} is the log-partition function of a canonical exponential family. Using \Cref{app: lem: kl divergence of two exponential family distributions}, and plugging $v = \xi_{\traindata}$, $u = \xi_{\traindata + \testdata}$, and $w = \xi_{\traindata + 2\testdata}$  for conjugate prior family gives \cref{eq: exp family delta form 1.}.
  \end{proof}

\newpage
\section{Proof for \Cref{thm: snr monte carlo under q.}}
\label{app: final sec: theorem approximate inference.}
\begin{defn}
  \label{app: def: log constant and posterior for approx}
  Let $p(\traindata \vert z)$ be the likelihood and $p(z)$ be the prior distribution. Let $q_{\traindata}(z)$ be the variational distribution. Let $\testdata$ be some testdata. Then, 
  \begin{align}
    \genvarlogconst(\testdata)
    &\coloneqq \log \int p(\testdata \vert z) q_{\traindata}(z) dz
    \quad \quad \quad \text{and} \quad \quad \quad 
    q_\traindata(z \vert \testdata) \coloneqq \frac{p(\testdata \vert z)q_{\traindata}(z)}{\genvarlogconst(\testdata)}
  \end{align}
\end{defn}

\begin{lem}
  \label{app: lem: moments of r1 approx}
  Let $p(\traindata \vert z)$ be the likelihood and $p(z)$ be the prior distribution. Let $q_{\traindata}(z)$ be the variational distribution. 
  Let $\testdata$ be some test data. Let $p(\traindata + \testdata \vert z) = p(\traindata \vert z)p(\testdata \vert z)$ for any datasets $\traindata$ and $\testdata$. Let $R_{K}$ be the Monte Carlo estimator for the $\PPDp$ under approximate inference (\cref{eq: naive mc estimator for ppdq.} with $K = 1$.) Then, 
  \begin{align}
    \E \left[R_{1}^c\right] = \exp \genvarlogconst(c\testdata),
  \end{align}
  where $c$ is a non-negative integer.
\end{lem}
\begin{proof}
  The proof is straightforward for $c = 0$ as $Z_\traindata(\emptyset) = \log \int q_\traindata(z) dz = 0$.
  For $c \ge 1$, we have
  \begin{align}
    \E \left[R_1^c\right] 
    &= \E \left[ p(\testdata \vert z)^c \right]\\
    &= \E \left[ p(c\testdata \vert z) \right]\\
    &= \int p(c\testdata \vert z) q_{\traindata}(z) dz\\
    &= \exp \genvarlogconst(c\testdata).
  \end{align}
\end{proof}

\begin{lem}
  \label{app: lem: kl divergence between augmented posteriors approx}
  Let $p(\traindata \vert z)$, $p(z)$, and $q_{\traindata}(z)$ be as in \cref{app: def: log constant and posterior for approx}. 
  Let $\traindata_a$ and $\traindata_b$ be the three multisets of data.
  Then,
  \begin{align}
    \KL{q_\traindata(z \vert \traindata_a)}{q_\traindata(z \vert \traindata_b)} 
    &= \E \left[\log p(\traindata_a \vert z) - \log p(\traindata_b \vert z)\right]  - \genvarlogconst(\traindata_a) + \genvarlogconst(\traindata_b)\\
  \end{align}
\end{lem}
\begin{proof}
  \begin{align}
    &\KL{q_\traindata(z \vert \traindata_a)}{q_\traindata(z \vert \traindata_b)} \\
    & = \E \left[\log \frac{\frac{p(\traindata_a \vert z)q_{\traindata}(z)}{\exp \genvarlogconst(\traindata_a)}}{\frac{p(\traindata_b \vert z)q_{\traindata}(z)}{\exp \genvarlogconst(\traindata_b)}}\right]\\
    & = \E \left[\log \frac{p(\traindata_a \vert z)}{p(\traindata_b \vert z)}\right]  - \log \frac{\exp \genvarlogconst(\traindata_a)}{\exp \genvarlogconst(\traindata_b)}\\
    & = \E \left[\log \frac{p(\traindata_a \vert z)}{p(\traindata_b \vert z)}\right]  - \genvarlogconst(\traindata_a) + \genvarlogconst(\traindata_b)\\
  \end{align}
\end{proof}

\begin{lem}
  \label{app: lem: sum of kl divergence approx}
  Let $p(\traindata \vert z)$, $p(z)$, and $q_{\traindata}(z)$ be as in \cref{app: def: log constant and posterior for approx}. Let $\traindata_1$, $\traindata_2$, and $\traindata_3$ be the three multisets of data.
  Let $\traindata_1$, $\traindata_2$, and $\traindata_3$ be the three multisets of data.
  Then,
  \begin{align}
    &\frac{1}{2}\KL{q_\traindata(z \vert \traindata_3)}{q_\traindata(z \vert \testdata_1)} 
    + \frac{1}{2}\KL{q_\traindata(z \vert \traindata_3)}{q_\traindata(z \vert \testdata_2)} \\
    & = \E \left[\log p(\traindata_3 \vert z) - \frac{1}{2}\log p(\traindata_1 \vert z) - \frac{1}{2}\log p(\traindata_2 \vert z)\right]  \\
    & + \frac{\genvarlogconst(\traindata_1)  + \genvarlogconst(\traindata_2)}{2} - \genvarlogconst(\traindata_3).
  \end{align}
\end{lem}
\begin{proof}
  Applying the \cref{app: lem: kl divergence between augmented posteriors approx} to $\traindata_3$ and $\traindata_1$ gives 
  \begin{align}
    &\KL{q_\traindata(z \vert \traindata_3)}{q_\traindata(z \vert \testdata_1)} \\
    & = \E \left[\log p(\traindata_3 \vert z) - \log p(\traindata_1 \vert z)\right]  - \genvarlogconst(\traindata_3) + \genvarlogconst(\traindata_1)\\
  \end{align}
  and applying it to $\traindata_3$ and $\traindata_2$ gives
  \begin{align}
    &\KL{q_\traindata(z \vert \traindata_3)}{q_\traindata(z \vert \testdata_2)} \\
    & = \E \left[\log p(\traindata_3 \vert z) - \log p(\traindata_2 \vert z)\right]  - \genvarlogconst(\traindata_3) + \genvarlogconst(\traindata_2).\\
  \end{align}
  Now, multiplying the above two equations by $\frac{1}{2}$ and adding them gives
  \begin{align}
    &\frac{1}{2} \KL{q_\traindata(z \vert \traindata_3)}{q_\traindata(z \vert \testdata_1)}\\
    & + \frac{1}{2} \KL{q_\traindata(z \vert \traindata_3)}{q_\traindata(z \vert \testdata_2)} \\
    & = 
    \E \left[\log p(\traindata_3 \vert z) - \frac{1}{2}\log p(\traindata_1 \vert z) - \frac{1}{2}\log p(\traindata_2 \vert z)\right]  \\
    & + \frac{\genvarlogconst(\traindata_1)  + \genvarlogconst(\traindata_2)}{2} - \genvarlogconst(\traindata_3).
  \end{align}
\end{proof}

\begin{cor}
  \label{app: cor: sum of kl divergence approx}
  Let $p(\traindata \vert z)$, $p(z)$, and $q_{\traindata}(z)$ be as in \cref{app: def: log constant and posterior for approx}. Let $\traindata_1 = c_a\traindata$, $\traindata_2 = c_a\traindata + 2c_b \testdata$, and $\traindata_3 = c_a\traindata + c_b\testdata$ be the three multisets of data where $c_a$ and $c_b$ are non-negative integers. Then,
  \begin{align}
    &\frac{1}{2}\KL{q_\traindata(z \vert \traindata_3)}{q_\traindata(z \vert \testdata_1)} 
    + \frac{1}{2}\KL{q_\traindata(z \vert \traindata_3)}{q_\traindata(z \vert \testdata_2)} \\
    & =\frac{\genvarlogconst(\traindata_1)  + \genvarlogconst(\traindata_2)}{2} - \genvarlogconst(\traindata_3).
  \end{align}
\end{cor}
\begin{thm}
  \label{app: thm: delta approx}
  Let $p(\traindata \vert z)$ be the likelihood and $p(z)$ be the prior distribution. Let $q_{\traindata}(z)$ be the variational distribution. 
  Let $\testdata$ be some testdata. 
  Let $p( \traindata + \testdata \vert z) = p(\traindata \vert z)p(\testdata \vert z)$ for any datasets $\traindata$ and $\testdata$. 
  Let $R_{K}$ be the Monte Carlo estimator for the $\PPDp$ under approximate inference (\cref{eq: naive mc estimator for ppdq.} with $K = 1$.) Then, the signal-to-noise ratio of $R_{K}$ is given by
  $\SNR{R_{K}} = \sqrt{K}/\sqrt{\exp(\delta)^2 - 1}$ where
  \begin{align}
    \delta 
    & = \frac{1}{2} \KL{q_\traindata(z \vert \testdata)}{q_{\traindata}(z)} + \frac{1}{2} \KL{q_\traindata(z \vert \testdata)}{q_\traindata ( z \vert 2\testdata)}\\
    & = \frac{1}{2}\genvarlogconst(2\testdata)  - \genvarlogconst(\testdata) 
  \end{align}
  where $\genvarlogconst$ and $q_\traindata(z \vert \testdata)$ are as in \cref{app: def: log constant and posterior for approx}.

\end{thm}

\begin{proof}
  \begin{align}
    \delta 
    &\overset{(a)}{=} \frac{1}{2} \log \frac{\E[R_{1}^2]}{\E[R_{1}]^2} \\
    &= \frac{1}{2} \log \E \left[R_{1}^2\right] - \log \E \left[R_{1}\right]\\
    &\overset{(b)}{=} \frac{1}{2}\genvarlogconst(2\testdata)  - \genvarlogconst(\testdata) 
  \end{align}
  Where $(a)$ follows from \cref{app: lem: snr of naive monte carlo estimator} and $(b)$ follows from \cref{app: lem: moments of r1 approx}.
  Lastly, plugging $\traindata_1 = \emptyset$ and $\traindata_2 = 2 \testdata$ and $\traindata_3 = \testdata$ into \cref{app: cor: sum of kl divergence approx} and observing $\genvarlogconst(\emptyset) = 0$ gives the result.
\end{proof}
\newpage

\section{Proof for \Cref{cor: snr monte carlo under q and conjugacy.}}
\label{app: final sec: cor. approximate inference.}

\begin{lem}
  \label{app: lem: r1 moments approx exp fam}
  Let the likelihood $p(y \vert z)$ be as in \cref{eq: conjugate likelihood} and a prior $p(z)=s(z \vert \xi_0)$ be as in \cref{ eq: normalization constant for conjugate prior}. Let $q_{\traindata}(z) = s(z \vert \eta)$ be in the conjugate family (\cref{ eq: normalization constant for conjugate prior}.) Let $\testdata$ be some test data and let $R_{1}$ be the Monte Carlo estimator for the $\PPDp$ under approximate inference (\cref{eq: naive mc estimator for ppdq.} with $K = 1$.) 
  Then, 
  \begin{align}
    \E[R_{1}^c] 
    &= h(\testdata)^c \exp\left(B\left(\varparamone + \cteststatthree{U}{c}{\testdata}\right) - B\left(\varparamone\right)\right),
  \end{align}
$c$ is a non-negative integer, 
$B$ is as in \cref{ eq: normalization constant for conjugate prior}, and $\cteststatthree{U}{c}{\traindata} = \cteststattwo{c}{T}{\traindata}$ for any dataset $\traindata$.
\end{lem}

\begin{proof}
  Starting from the definition of $R_{q,1}$ we have,
  \begin{align}
    \E[R_{q,1}^c] 
    = \E\left[\left(p(\testdata\vert z)\right)^c\right]
    &
    =\E\left[\left(\prod_{y \in \testdata} p(y \vert z)\right)^c\right]\\
    &=\E\left[\left(\prod_{y \in \testdata} h(y) \exp\left(T(y)^\top \phi(z) - A(z)\right)\right)^c\right]\\
    &\overset{\textrm{(a)}}{=} \E\left[\left(h(\testdata) \exp\left(T(\testdata)^\top \phi(z)  - |\testdata|A(z)\right)\right)^c\right],
  \end{align}
  where $\text{(a)}$ follows from $T(\testdata) = \sum_{y \in \testdata} T(y)$ and $h(\testdata) = \prod_{y\in \testdata} h(y)$. Doing some basic manipulations, we get
  \begin{align}
    &\E\left[\left(h(\testdata) \exp\left(T(\testdata)^\top \phi(z)  - |\testdata|A(z)\right)\right)^c\right]\\
    &= h(\testdata)^c \E\left[\exp\left(cT(\testdata)^\top \phi(z)  - c|\testdata|A(z)\right)\right]\\
    &\overset{\textrm{(b)}}{=} h(\testdata)^c \E \left[\exp\left( c\left(\begin{bmatrix}
      T(\testdata)\\
      |\testdata|
    \end{bmatrix}\right)^\top\begin{bmatrix}\phi(z) \\ -A(z)\end{bmatrix}\right)\right]\\
    &\overset{\textrm{(c)}}{=} h(\testdata)^c \E \left[\exp\left( \cteststatthree{U}{c}{\testdata}^\top\begin{bmatrix}\phi(z) \\ -A(z)\end{bmatrix}\right)\right]\\
    &\overset{\textrm{(d)}}{=} h(\testdata)^c \int \exp\left( \cteststatthree{U}{c}{\testdata}^\top\begin{bmatrix}\phi(z) \\ -A(z)\end{bmatrix}\right) s(z \vert \varparamone) dz\\
    &\overset{\textrm{(e)}}{=} h(\testdata)^c \frac{\int \exp\left(\cteststatthree{U}{c}{\testdata}^\top\begin{bmatrix}\phi(z) \\ -A(z)\end{bmatrix}\right)  \exp\left( \varparamone^\top\begin{bmatrix}\phi(z) \\ -A(z)\end{bmatrix}\right) dz}{\exp B(\varparamone)}\\
    &\overset{\textrm{(f)}}{=} h(\testdata)^c \frac{\int \exp\left( \left(\cteststatthree{U}{c}{\testdata} + \eta\right)^\top\begin{bmatrix}\phi(z) \\ -A(z)\end{bmatrix}\right)   dz}{\exp B(\varparamone)}\\
    &\overset{\textrm{(g)}}{=} h(\testdata)^c \frac{\exp(B(\varparamone + \cteststatthree{U}{c}{\testdata}))}{\exp(B(\varparamone))}\\
    &= h(\testdata)^c \exp(B(\varparamone + \cteststatthree{U}{c}{\testdata}) - B(\varparamone))
  \end{align}
  where
  (b) just collects the terms in the exponent into a single vector;
  (c) defines $\cteststatthree{U}{c}{\traindata} = \cteststattwo{c}{T}{\traindata}$ for any dataset $\traindata$;
  (d) and (e) follows as expectation is under the variational distribution and the definition of conjugate family in \cref{ eq: normalization constant for conjugate prior}; 
  (f) follows from some simple algebra;
  (g) follows from the definition of $B$ in \cref{ eq: normalization constant for conjugate prior}.
\end{proof}

\begin{thm}\label{app: thm: ppdq snr monte carlo}
Take a model with a likelihood $p(y \vert z)$ in an exponential family (\cref{eq: conjugate likelihood}) and a prior $p(z)=s(z \vert \xi_0)$ in the corresponding conjugate family (\cref{ eq: normalization constant for conjugate prior}). Let $q_{\traindata}(z) = s(z \vert \varparamone)$ be an approximate distribution in the corresponding conjugate family (\cref{ eq: normalization constant for conjugate prior}) with parameters $\varparamone$.
Let $\testdata$ be a multiset of test data and let $R_{1, q}$ be the Monte Carlo estimator for the $\PPDp$ (\cref{eq: naive mc estimator for ppdq.} with $K = 1$.) 
Then, the signal-to-noise ratio is $\text{SNR} (R_{1, q}) = \frac{1}{ \sqrt{\exp\left(\delta\right)^2 - 1}}$ for
\begin{align}
\delta 
  &=\frac{1}{2} \KL{s\pp{z\vert \varparamone + \cteststatthree{U}{}{\testdata}}}{s\pp{z\vert \varparamone}} + \frac{1}{2} \KL{s\pp{z\vert \varparamone + \cteststatthree{U}{}{\testdata}}}{s\pp{z\vert \varparamone + \cteststatthree{U}{2}{\testdata}}} \label{app: eq: delta for ppdq monte carlo KL} \\ 
   &= \frac{B\left(\varparamone\right) + B(\varparamone + \cteststatthree{U}{2}{\testdata})}{2}  - B\left(\varparamone + \cteststatthree{U}{}{\testdata}\right), \label{app: eq: delta for ppdq monte carlo b}
\end{align}
where $B$ is as in \cref{ eq: normalization constant for conjugate prior} and
$\cteststatthree{U}{c}{\traindata} = \cteststattwo{c}{T}{\traindata}$ for any dataset $\traindata$ and non-negative integer $c$.
\end{thm}

\begin{proof}
  From \Cref{app: lem: snr of naive monte carlo estimator} we get $\SNR{R_{K}} = \frac{\sqrt{K}}{\sqrt{\exp(\delta)^2 - 1}}$ for $\delta = \frac{1}{2} \log(\E[R_{1}^2]/\E[R_{1}]^2)$. Then
  \begin{align}
    \delta 
    = \frac{1}{2} \log \frac{\E[R_{1}^2]}{\E[R_{1}]^2} 
    &= \frac{1}{2} \log \E\left[R_{1}^2\right] - \log \E\left[R_{1}\right]\\
    &\overset{\textrm{(a)}}{=} \frac{1}{2}\left(B(\varparamone + \cteststatthree{U}{2}{\testdata}) - B(\varparamone)\right) - \left(B(\varparamone + \cteststatthree{U}{}{\testdata}) - B(\varparamone)\right)\\
    &\overset{\textrm{(b)}}{=} \frac{B(\varparamone + \cteststatthree{U}{2}{\testdata}) + B(\varparamone)}{2} - B(\varparamone + \cteststatthree{U}{}{\testdata}),
  \end{align}
  where (a) follows from \Cref{app: lem: r1 moments approx exp fam} for $c =1$ and $c = 2$ and cancellations of $\log h(\testdata)$ terms and (b) form simple algebraic manipulations.
  
  Now, observe $B$ in \cref{ eq: normalization constant for conjugate prior} is the log-partition function of a canonical exponential family. Using \Cref{app: lem: kl divergence of two exponential family distributions}, and plugging $v = \varparamone$, $u = \varparamone + \cteststatthree{U}{}{\testdata}$, and $w = \varparamone + \cteststatthree{U}{2}{\testdata}$  for conjugate prior family gives the \cref{eq: exp fam q delta form 1}.
\end{proof}


\newpage
\section{General experimental details}
\label{app: final sec: general experimental details.}

All our code is implemented in JAX \citep{jax2018github} and run on a single NVIDIA A100 GPU. 
In \Cref{tab: expressions and computations v2}, we provide the expressions for computation of different metrics from the results in \Cref{tab: ppd estimation,tab: approx ppd estimation,tab: log reg ppd estimation,subsec: movielens 25m.}. 

\textbf{General Note on BBVI:} We rely on using standard BBVI techniques for most of our experiments. The hope of BBVI is to allow practitioners to not worry about designing special approximation families for each model $p(\traindata,z)$ \citep{ranganath14,kucukelbir2017automatic,aagrawal2020,ambrogioni2021automatic,ambrogioni2021automaticb,burroni2023sample}. Instead, BBVI treats models as black boxes\textemdash only requiring access to $\nabla_z \log p(\traindata, z)$ to update the variational parameters using the stochastic gradients of a variational objective (for instance, $\IWELBO$.) 
Ongoing research in BBVI focuses on automating other algorithmic choices \citep{kucukelbir2017automatic,aagrawal2020,ambrogioni2021automatic,ambrogioni2021automaticb,burroni2023sample}. Such optimization schemes greatly improve the applicability of BBVI and come pre-implemented in popular probabilistic programming languages like Pyro \citep{bingham2019pyro}, NumPyro \citep{phan2019composable}, and Stan \citep{carpenter2017stan}. While we implement our own inference schemes for this paper, we expect the results to be similar if we use the aforementioned libraries.


\begin{table}[H]
  \caption{
    \label{tab: expressions and computations v2} Summary of the expressions of metrics and their computations for the table \Cref{tab: ppd estimation,tab: approx ppd estimation,tab: log reg ppd estimation,tab: movielens 25m ppd estimation}. We report $\yhwidehat{\SNR{R}}$ in terms of $\yhwidehat{\E[R]}$ and $\yhwidehat{\V[R]}$ and report explicit form in \Cref{tab: mean for snr,tab: variance for snr}. We use $S = 1000$ for all our experiments. The results are then averaged over ten independent trials to generate mean and standard deviation numbers in \Cref{tab: approx ppd estimation,tab: log reg ppd estimation,tab: movielens 25m ppd estimation,tab: ppd estimation}
  }
  \begin{center}
    \resizebox{\textwidth}{!}{
      \begin{tabular}{@{}ll@{\hspace{40pt}}ll@{}}
        \toprule
        Expression & Computation & Expression & Computation\\
        \midrule
        $\E[{\log R_{K}}]$ & $z_{s,k} \sim q(z\vert \traindata), \frac{1}{S}\sum_{s=1}^{S}\left[\log\frac{1}{K}\sum_{k=1}^{K} p(\testdata \vert z_{s, k})\right]$ & $\yhwidehat{\SNR{R_{K}}}$ & ${\yhwidehat{\E [{R_{K}}]}}\bigg/{\sqrt{\yhwidehat{\V [{R_{K}}]}}}$ \\[1.5em]
        $\E[{\log R_{K}^{\mathrm{IS}}}]$ &$z_{s,k} \sim r_w(z), \frac{1}{S}\sum_{s=1}^{S}\left[\log\frac{1}{K}\sum_{k=1}^{K} \frac{p(\testdata \vert z_{s, k})q(z_{s, k} \vert \traindata)}{r_w(z_{s, k})}\right]$ & $\yhwidehat{\SNR{R_{K}^{\mathrm{IS}}}}$ &${\yhwidehat{\E [{R_{K}^{\mathrm{IS}}}]}}\bigg/{\sqrt{\yhwidehat{\V [{R_{K}^{\mathrm{IS}}}]}}}$ \\[1.5em]
        \bottomrule
      \end{tabular}
      }

    \end{center}
\end{table}

  \begin{table}[H]
    \caption{\label{tab: mean for snr} \small{Mean of SNR for different estimators.}}
    \begin{center}
      \resizebox{0.6\linewidth}{!}{
        \begin{tabular}{@{}ll@{}}
          \toprule
          Expression & Computation\\
          \midrule
          $\yhwidehat{\E [{R_{K}}]}$ & $z_{s,k} \sim q(z\vert \traindata), \frac{1}{S} \sum_{s = 1}^{S}  \left[\frac{1}{K} \sum_{k=1}^{K}p(\testdata \vert z_{s, k})\right] $\\[2em]
          $\yhwidehat{\E [{R_{K}^{\mathrm{IS}}}]}$ & $z_{s,k} \sim r_w(z), \frac{1}{S} \sum_{s = 1}^{S}  \left[\frac{1}{K} \sum_{k=1}^{K}\frac{p(\testdata \vert z_{s, k})q(z_{s, k} \vert \traindata)}{r_w(z_{s, k})}\right] $\\[2em]
          \bottomrule
        \end{tabular}
        }
  
      \end{center}
  \end{table}
  
  \begin{table}[H]
    \caption{\label{tab: variance for snr} \small{Variance of SNR for different estimators.}}
    \begin{center}
      \resizebox{0.8\linewidth}{!}{
        \begin{tabular}{@{}ll@{}}
          \toprule
          Expression & Computation\\
          \midrule
          $\yhwidehat{\V [{R_{K}}]}$ & $z_{s,k} \sim q(z\vert \traindata), \frac{1}{S-1} \sum_{s = 1}^{S} \left[\frac{1}{K} \sum_{k=1}^{K}p(\testdata \vert z_{s, k}) - \yhwidehat{\E[R_{K}]} \right]^2$\\[2em]
          $\yhwidehat{\V [{R_{K}^{\mathrm{IS}}}]}$ & $z_{s,k} \sim r_w(z), \frac{1}{S-1} \sum_{s = 1}^{S} \left[\frac{1}{K} \sum_{k=1}^{K}\frac{p(\testdata \vert z_{s, k})q(z_{s, k} \vert \traindata)}{r_w(z_{s, k})} - \yhwidehat{\E[R_{K}^{\mathrm{IS}}]} \right]^2$\\[2em]
          \bottomrule
        \end{tabular}
        }
  
      \end{center}
  \end{table}


\newpage
\section{Exponential Family models: Additional Details}
\label{app: final sec: exponential family.}
For each of the three models, we fix the number of training data points $|\traindata| = 100$ and number of test data points $|\testdata| = 100$. 
Then, to sample the training data such that the mean statistics of the data $\overline{T(\traindata)} \approx 10$, we sample from the likelihood distributions by carefully adjusting the parameters. This means, for normal we sample from $\mathcal{N}(10, 1)$; for Exp we sample from $\mathrm{Exp}(0.1)$; and for Binomial we sample from $\mathrm{Binomial}(100, 0.1)$.

Then, to sample the test data, we first select the region of high $\delta$ from the \Cref{fig: snr for exp models} and then roughly try to match the target mean statistics by carefully adjusting the parameters. For Normal, we sample from $\mathcal{N}(5, 1)$ to target $\overline{T(\testdata)} \approx 5$; for Exp Ze sample from $\mathrm{Exp}(0.025)$ to target $\overline{T(\testdata)} \approx 40$; and for Binomial we sample from $\mathrm{Binomial}(100, 0.4)$ to target $\overline{T(\testdata)} \approx 40$. This strategy leads to the numbers in \Cref{tab: statistics of the data}. Note, we only use one test and train setting for our experiments. The results reported in \Cref{tab: ppd estimation,tab: approx ppd estimation} are averaged our ten independent estimations for a single data setting.

\begin{table}[H]
    \begin{center}
      \caption{\label{tab: conjugate models.} For the three models: Normal, Exp, and Binomial, we identify the exponential family form from \Cref{sec: conjugate model analysis}. For likelihood in \cref{eq: conjugate likelihood}, we identify base measure $h(y)$, one-to-one parameter mapping $\phi(z)$, and log-partition function $A(z)$. Note, the sufficient statistics $T(y) = y$ for all models. For the conjugate prior in \cref{ eq: normalization constant for conjugate prior}, we identify the log partition function $B(\xi)$, where $\xi = (\xi_T, \xi_n)^\top$. \vspace{10pt}}
      \resizebox{0.9\textwidth}{!}{
        \begin{tabular}[t]{@{}llllll@{}}
          \toprule
          Model &$p(y\vert z)$&$h(y)$  & $\phi(z)$ & $A(z)$& $B(\xi)$\\
          \midrule
          Normal  & $\mathcal{N}(y \vert z, \sigma^2)$ &$\frac{\exp(-\frac{y^2}{2 \sigma^2})}{\sqrt{2\pi \sigma^2}}$  & $\frac{z}{\sigma^2}$ & $\frac{z^2}{2\sigma^2}$ & $\frac{1}{2}\left[\log \frac{2 \pi \sigma^2}{\xi_n} +  \frac{\xi_T^2}{\sigma^2 \xi_n}\right]$\\[1.em]
          Exp & $\mathrm{Exp} (y \vert z)$ & $1$  & $-z$ & $-\log z$ & $ \log \frac{\Gamma(\xi_n + 1)}{\xi_T^{\xi_n + 1}}$\\[1.em]
          Binomial & $\mathrm{Bin}(y \vert n, z)$ & $\binom{n}{y}$  & $\log \frac{z}{1-z}$ & $-n\log (1-z)$& $\log  \frac{\Gamma(\xi_T + 1)\Gamma (n\xi_n - \xi_T + 1)}{\Gamma (n\xi_n + 2)}  $\\
          \bottomrule
        \end{tabular}
        }
      \end{center}
  \end{table}
  We learn a Gaussian variational approximation for each of the three models from \Cref{tab: approx ppd estimation}. For the models with constrained latent variables (Exponential and Binomial,) we transform $z$ to an unconstrained space and then adjust the logarithm of the determinant of the jacobian for correct density evaluation (please, see \cite[Section 2.3]{kucukelbir2017automatic} for more details on such transformations.) 
  Our variational family has two unconstrained parameters: $\mu$ and $\sigma$. 
  To ensure positivity of standard deviation, we transform $\sigma$ with the soft-plus function.

  We consider two options to initialize $\mu$ and $\sigma$: Laplace's approximation and standard Normal. 
  To pick from the two options, we evaluate ELBO using $1000$ samples and chose the option with higher ELBO value. 
  For Laplace's approximation, we use JAX's BFGS optimizer \citep{jax2018github} (for each model, BFGS took less than 50 evaluations of $\log p(z, \traindata)$.)
  
  To learn the variational parameters, we optimize standard ELBO using ADAM \citep{kingma2014ADAM} with a learning rate of $0.001$ for $10,000$ iterations. 
  For each iteration, we use a batch of 16 samples for estimating the DReG gradient \citep{tucker2018doubly}.
  
  We learn a parameterized Gaussian proposal distribution for each of the three models from \Cref{tab: ppd estimation,tab: approx ppd estimation}. For the models with constrained latent variables (Exponential and Binomial,) we transform $z$ to an unconstrained space and then adjust the logarithm of the determinant of the jacobian for correct density evaluation (please, see \cite[Section 2.3]{kucukelbir2017automatic} for more details on such transformations.) Our parameterized proposal distribution has two unconstrained parameters: $\mu$ and $\sigma$. To ensure positivity of standard deviation, we transform $\sigma$ with the soft-plus function.
  
  We consider two options to initialize $\mu$ and $\sigma$: Laplace's approximation and standard Normal. To pick from the two options, we evaluate $\IWELBO_M$ using $1000$ samples and chose the option with higher $\IWELBO_M$ value. For Laplace's approximation, we use JAX's BFGS optimizer \citep{jax2018github} (for each model, BFGS took less than 50 evaluations of $\log p(\testdata \vert z)p(z \vert \traindata)$ or $p(\testdata \vert z) q_{\traindata}(z)$.)
  
  To learn the proposal parameters, we optimize $\IWELBO_M$ using ADAM \citep{kingma2014ADAM} with a learning rate of $0.001$ for $1000$ iterations. For each iteration, we use a single sample of the DReG estimator. Note, a single sample of DReG estimator for $\IWELBO_M$ uses $M$ samples. We set $M = 16$ for all our experiments. Note, even after counting the Laplace's approximation evaluations, we use less than $20,000$ evaluations of $\log p(\testdata \vert z) p(z \vert \traindata)$ for learning the proposal.
  
  \begin{wrapfigure}{r}{0.4\textwidth}
      \centering
      \includegraphics[width=0.8\linewidth]{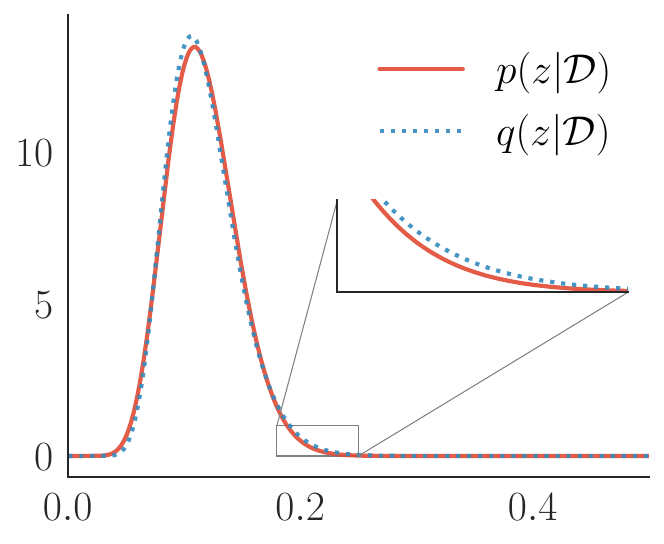}
      \caption{Fatter right tail of $q_\traindata$ for Binomial model.}
      \label{fig: binomial fatter right tail}
  \end{wrapfigure}
\paragraph*{Fatter tails.} For the Binomial model, we observe that the estimates for $\PPDp$ are higher than the estimates for $\PPDp$. This can be explained from two observations. First, the approximate posterior has fatter right tails than the true posterior, and second, the test data mean lies to the right of the training data mean (see \Cref{tab: statistics of the data}). This means that the approximate posterior places more mass in the region of test data and the $\PPDp$ will be higher than $\PPDp$. In \Cref{fig: binomial fatter right tail}, we plot the densities for the exact posterior and the learned approximation $q_\traindata$. We also plot an inset-zoomed-in version to highlight the fatter right tail of the approximate posterior. Remember, the variational approximation in the constrained space is obtained after transforming the unconstrained Gaussian variational approximation.

\subsection{Empirical Validation for \Cref{prop: delta Bayesian CLT}}
\label{app: subsec: visualizations for delta Bayesian CLT}
\begin{wrapfigure}{r}{0.45\textwidth}
  \vspace{-15pt}
  \centering
  \includegraphics[width=0.8\linewidth]{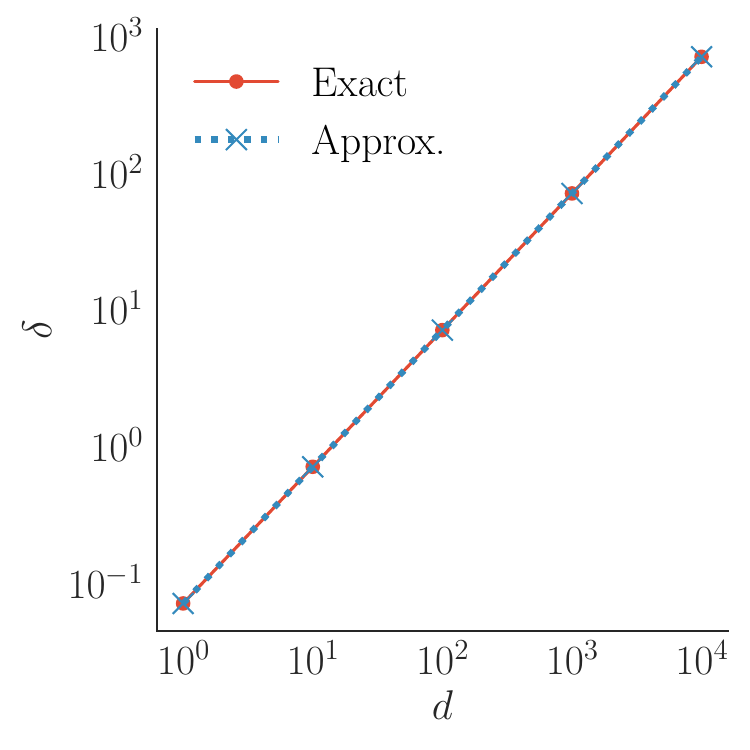}
  \caption{
    \small
    \label{fig: delta vs d} $\delta$ from approximation in \Cref{prop: delta Bayesian CLT} (blue dotted line) is accurate when compared to $\delta$ from exact expression in \cref{eq: new delta form 1.} (red solid lines.) Also, $\delta$ scales linearly with $d$ (\Cref{prop: delta Bayesian CLT}.) 
    }
  \end{wrapfigure}
We consider a model similar to the Normal model where likelihood $p(y \vert z)$ is given by a multivariate normal $\mathcal{N}(y \vert z, \Sigma)$ with unknown mean $z \in \R^d$ and known variance $\Sigma = \mathbb I_d$. A multivariate Normal prior $\mathcal{N}(z \vert 0, \mathbb I_d)$ gives a conjugate model as in \Cref{sec: conjugate model analysis}. 
For this model, we vary the number of latent dimensions $d \in \{1, 10, 100, 10000, 10000\}$. For each $d$, we create a training data set $\traindata$ with $1000$ data points, and set test data $\testdata$ to $\traindata$, that is, the mean statistics for training and test data sets match exactly. 
In \Cref{fig: delta vs d}, we plot the $\delta$ from the approximation in \Cref{prop: delta Bayesian CLT,eq: delta CLT} (shown in blue dotted lines with crosses), and compare it against the $\delta$ from exact calculations in  \cref{eq: new delta form 1.} (shown in red solid lines with dots). 
The approximation is accurate for all $d$, and $\delta$ scales linearly as predicted. This means for higher dimensional latent spaces, we can have extremely low $\SNR{R_1}$ even if test data statistics match exactly to the training data statistics.

\subsection{Figures for $\delta$ and $\SNR{R_1}$ contours}

\begin{figure}[h]
  \begin{subfigure}{0.33\textwidth}
    \includegraphics[width = \textwidth]{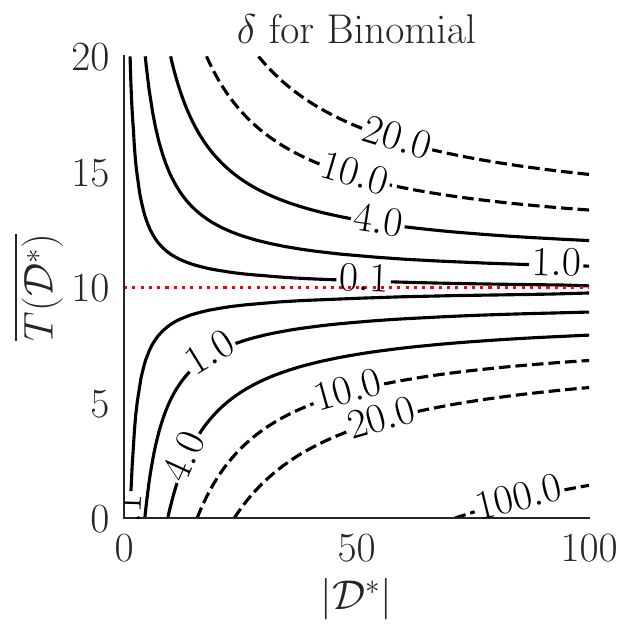}
  \end{subfigure}
  \begin{subfigure}{0.33\textwidth}
    \includegraphics[width = \textwidth]{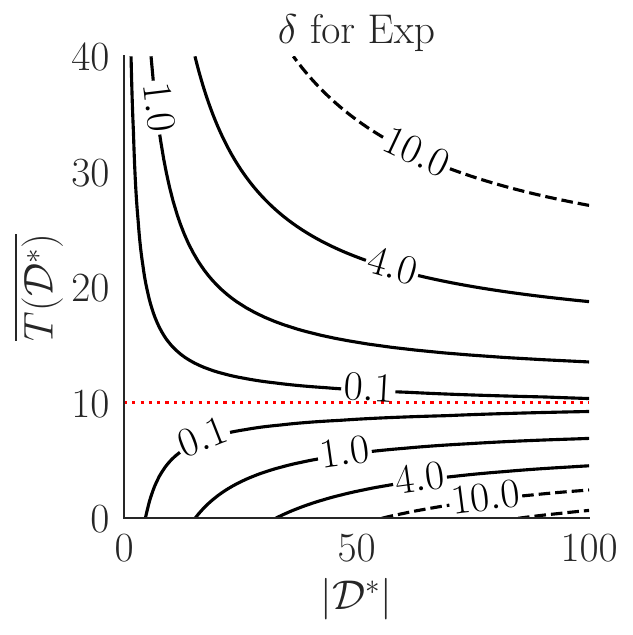}
  \end{subfigure}
  \begin{subfigure}{0.33\textwidth}
    \includegraphics[width = \textwidth]{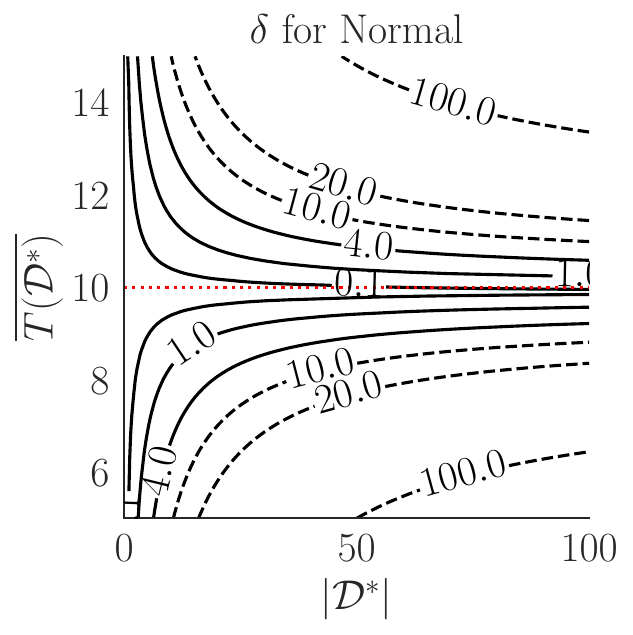}
  \end{subfigure}
  \caption{\label{fig: delta for different models repeated.} \textbf{$\delta$ contours.} Setting exactly the same as \Cref{fig: snr for exp models.}
  }
\end{figure}
\begin{figure}[h]
  \begin{subfigure}{0.33\textwidth}
    \includegraphics[width = \textwidth]{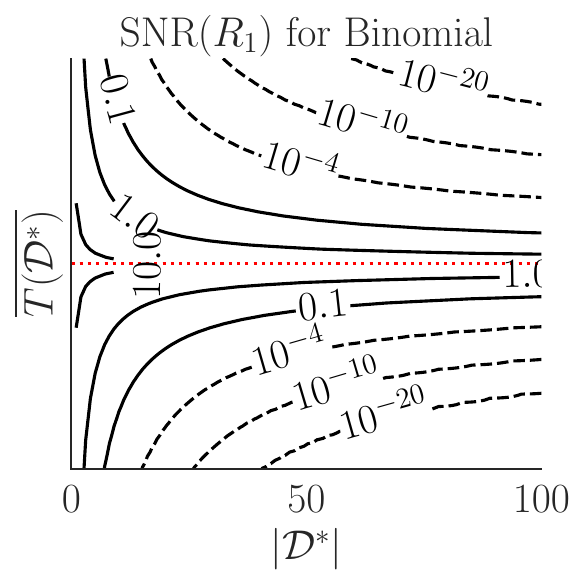}
  \end{subfigure}
  \begin{subfigure}{0.33\textwidth}
    \includegraphics[width = \textwidth]{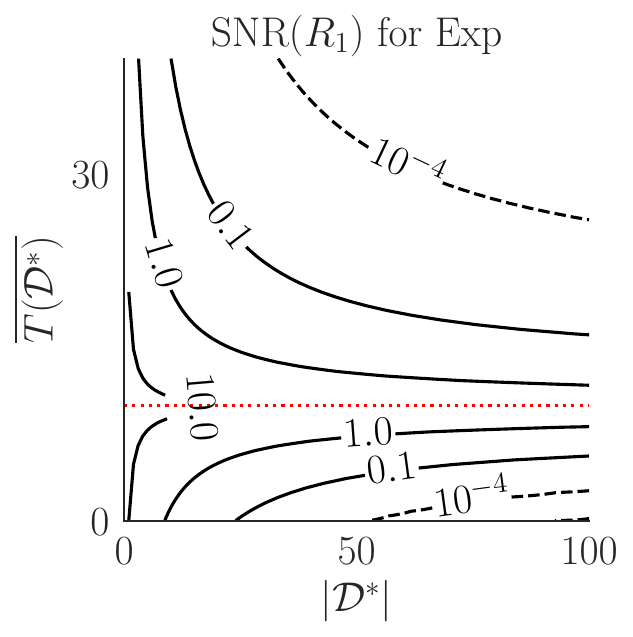}
  \end{subfigure}
  \begin{subfigure}{0.33\textwidth}
    \includegraphics[width = \textwidth]{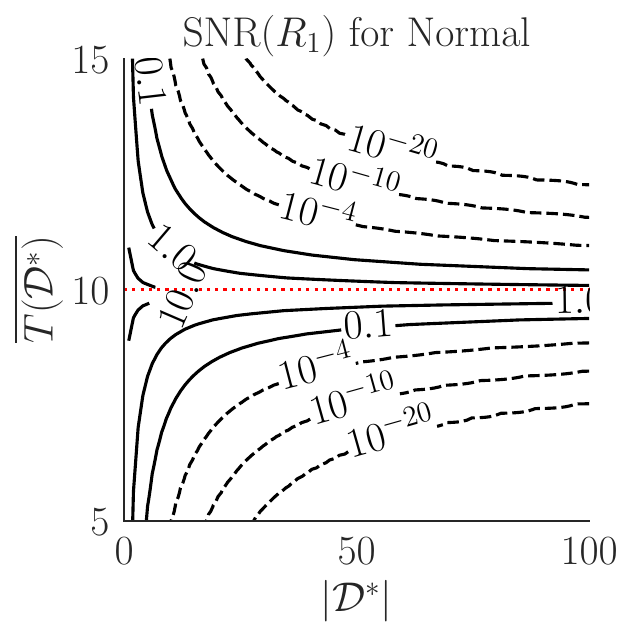}
  \end{subfigure}
  \caption{\label{fig: snr for different models repeated.} \textbf{SNR contours.} (Repeated for easier reference.) Setting is the same as the \Cref{fig: snr for exp models.}. 
  }
\end{figure}

\subsection{Effect of increasing the number of training data points}
\label{app: subsec: effect of increasing the number of training data points.}
\begin{figure}[h]
  \begin{subfigure}{0.33\textwidth}
    \includegraphics[width = \textwidth]{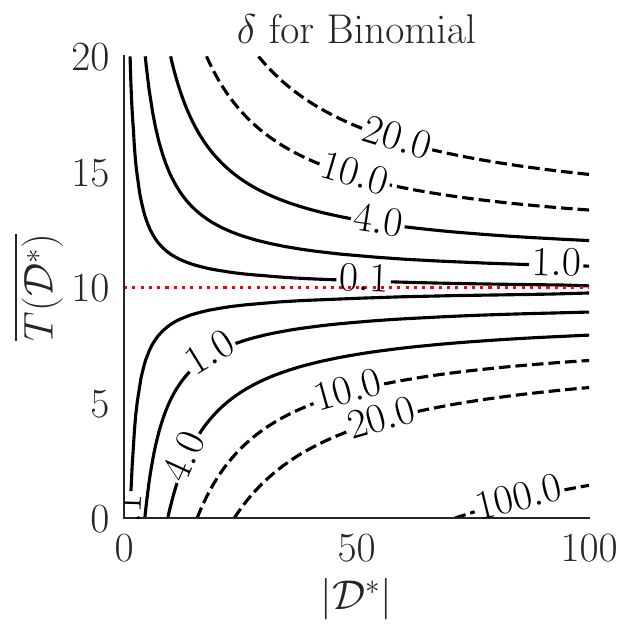}
  \end{subfigure}
  \begin{subfigure}{0.33\textwidth}
    \includegraphics[width = \textwidth]{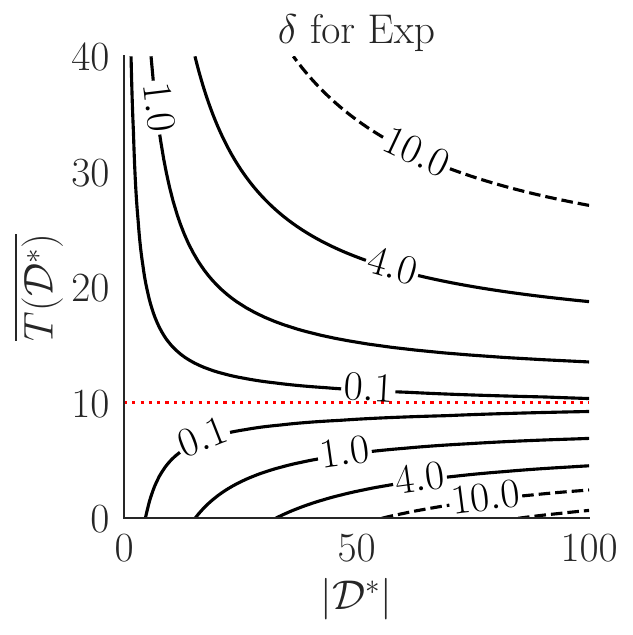}
  \end{subfigure}
  \begin{subfigure}{0.33\textwidth}
    \includegraphics[width = \textwidth]{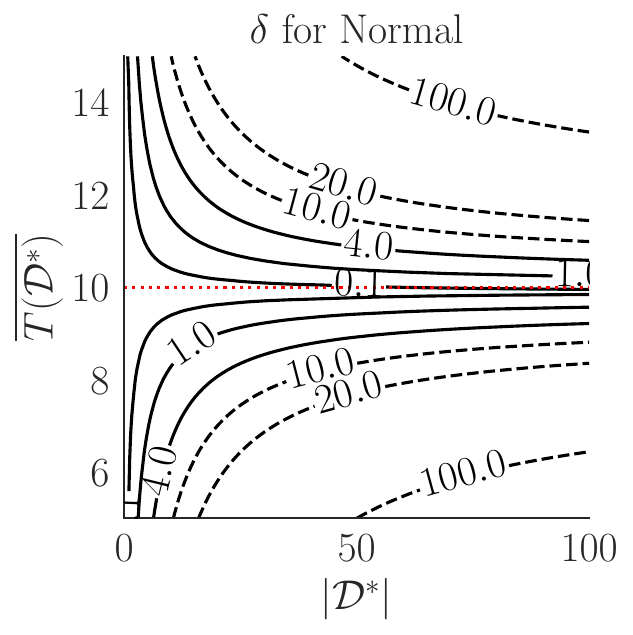}
  \end{subfigure}
  \caption{\label{fig: snr for different models repeated again.} (Repeated for easier reference.) $\mathbf{\delta}$ \textbf{contours.} Settings exactly the same as \Cref{fig: snr for exp models.}.
  }
\end{figure}

\begin{figure}[h]
  \begin{subfigure}{0.33\textwidth}
    \includegraphics[width = \textwidth]{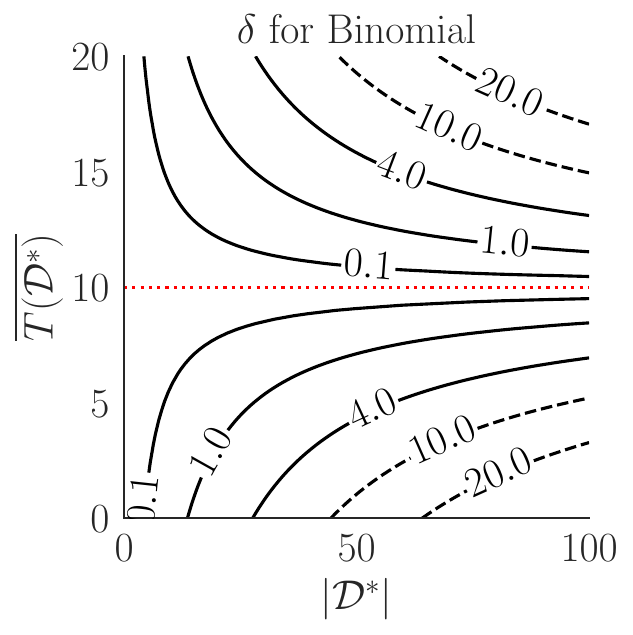}
  \end{subfigure}
  \begin{subfigure}{0.33\textwidth}
    \includegraphics[width = \textwidth]{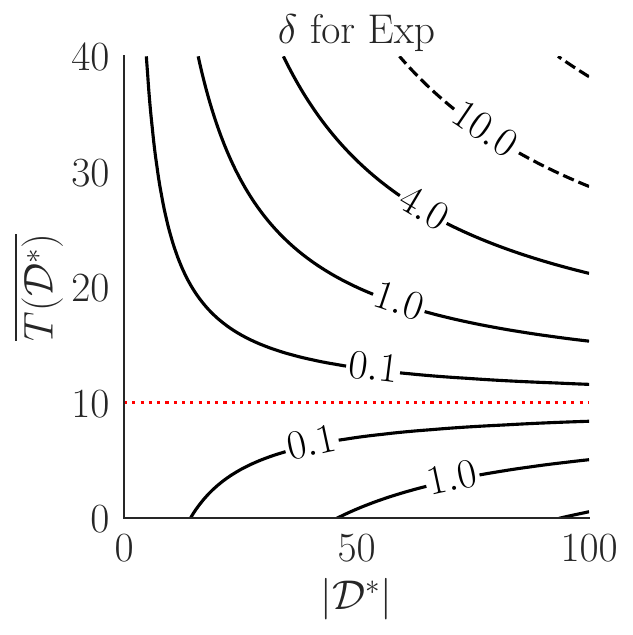}
  \end{subfigure}
  \begin{subfigure}{0.33\textwidth}
    \includegraphics[width = \textwidth]{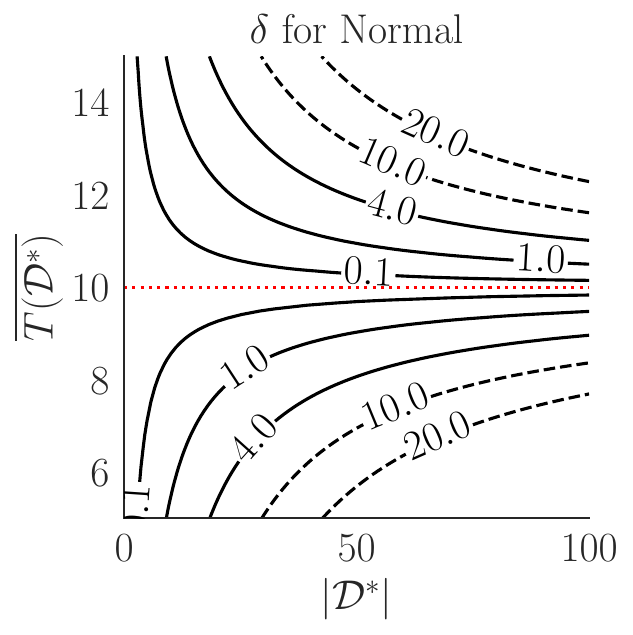}
  \end{subfigure}
  \caption{
    \label{fig: snr for different models with more training data.} $\mathbf{\delta}$ \textbf{contours.} For each model, we first fix the training data set such that $\overline{T(\traindata)} = 10$ (shown with red dotted line) and $|\traindata| = 1000.$   For all the models, increasing the number of training data points results in lower $\delta$ for a given test data statistics when compared to \Cref{fig: snr for different models repeated again.}. 
  }
\end{figure}

In \Cref{fig: snr for different models with more training data.}, we consider the effect of increasing the number of training data points from $|\traindata| = 100$ (\Cref{fig: snr for different models repeated again.},) to $|\traindata| = 1000$ while holding the mean training statistics, $\overline{T(\traindata)} = 10$, the same. 
As the number of training data points increases, $\delta$ gets smaller for any given test setting. 
To understand why, note $\delta$ as in \cref{eq: new delta form 1.} involves two KL divergences: one between posteriors $p(z \vert \traindata + \testdata)$ and $p(z \vert \traindata)$, and the other between posteriors $p(z \vert \traindata + \testdata)$ and $p(z \vert \traindata + 2\testdata)$. Intuitively, as the number of training data points increases, we either require more test data or bigger mismatch between test data and training data for the two KL divergences to be large. Thus, for any given test data setting, we expect $\delta$ to be smaller as the number of training data points increases.



\newpage
\section{Linear Regression: Additional Details}
\label{app: final sec: linear regression.}

\begin{defn}[Bayesian Linear Regression  Model]
  \label{app: defn: linear regression model}
  Consider the linear regression model with a Gaussian likelihood such that
  \begin{align}
    p(y_\traindata \vert z) = \mathcal{N}(y_\traindata \vert X_\traindata z, \sigma^2 I).
  \end{align}
  where $y_\traindata \in \R^{\verts{\traindata}}$ is the response vector, $X_\traindata \in \R^{\verts{\traindata}\times d}$ is feature matrix, and $\sigma^2$ is the variance. 

  The conjugate prior is a Gaussian distribution such that 
  \begin{align}
    p(z) = \mathcal{N}(z \vert \mu_0, \Sigma_0)
  \end{align}
  where $\mu_0$ is the mean and $\Sigma_0$ is the covariance. 
  Then, the posterior distribution is given by
  \begin{align}
    p(z \vert y_\traindata)
    & = \mathcal{N} (z \vert \mu_{\traindata}, \Sigma_{\traindata}),
  \end{align}
  where 
  \begin{align}
    \Sigma_{\traindata} 
     = \left(\frac{1}{\sigma^2} X_{\traindata}^\top X_{\traindata}   + \Sigma_0^{-1}\right)^{-1} \quad \quad  \text{and} \quad \quad 
    \mu_{\traindata} 
     = \Sigma_{\traindata}\left(\frac{1}{\sigma^2} X_{\traindata}^\top y_{\traindata} + \Sigma_0^{-1}\mu_0\right).
  \end{align}
\end{defn}

\begin{assump}
  \label{app: assump: vanishing prior}
  Let $y_\traindata$ be the training response vector and let $X_\traindata$ be the training feature matrix.
  Let the prior $p(z) = \mathcal{N}(z \vert \mu_0, \Sigma_0).$ Let $\verts{\mu_0} < \infty$ and
  let $X_\traindata$ and $\Sigma_0$ be such that $\left(X^\top_\traindata X_\traindata\right)^{-1}\Sigma_0^{-1} \approx \mathbf{0}.$
\end{assump}

\begin{assump}
  \label{app: assump: test data}
  Let $y_\traindata$ be the training response vector and let $X_\traindata$ be the training feature matrix.
  Let $y_\testdata$ be the test response vector such that $y_\testdata = y_\traindata + \Delta$. Let $X_\testdata$ be the test feature matrix such that $X_\testdata = X_\traindata$.
\end{assump}

\begin{lem}
  \label{app: lem: lin reg posterior with test data}
  Let $p$ be the Bayesian linear regression model from \cref{app: defn: linear regression model}.
  Let \Cref{app: assump: test data,app: assump: vanishing prior} hold.
  Let $c$ be a non-negative integer.
  Then,
  \begin{align}
    p(z \vert \traindata + c\testdata) = \mathcal{N}(z \vert \mu_{\traindata + c \testdata}, \Sigma_{\traindata + c\testdata}),
  \end{align}
  where   
  \begin{align}
    \Sigma_{\traindata + c\testdata} \approx \frac{1}{c+1} \left(\frac{1}{\sigma^2}X_\traindata^\top X_\traindata\right)^{-1}
    \quad \quad \text{and} \quad \quad
    \mu_{\traindata + c\testdata} \approx X_\traindata^{+} \left(y_\traindata + \frac{c}{c+1}\Delta\right)
  \end{align}
  where $X_\traindata^{+}$ is the pseudo-inverse such that $X_\traindata^{+} = \left(X_\traindata^\top X_\traindata\right)^{-1} X_\traindata^\top$.
\end{lem}
\begin{proof}
  We first massage the expressions for the covariance and the mean of the posterior distribution such that we can use the \cref{app: assump: test data,app: assump: vanishing prior}.
  \begin{align}
    &\Sigma_{\traindata}\\
    & = \left(\frac{1}{\sigma^2} X_{\traindata}^\top X_{\traindata}   + \Sigma_0^{-1}\right)^{-1}\\
    & = \left(\left(X_{\traindata}^\top X_{\traindata}\right)\left(\frac{1}{\sigma^2}I    + \left(X_{\traindata}^\top X_{\traindata}\right)^{-1}\Sigma_0^{-1}\right)\right)^{-1} \label{app: eq: post cov}
  \end{align}

  \begin{align}
    &\mu_{\traindata}\\
    & = \Sigma_{\traindata}\left(\frac{1}{\sigma^2} X_{\traindata}^\top y_{\traindata} + \Sigma_0^{-1}\mu_0\right)\\
    & = \left(\frac{1}{\sigma^2} X_{\traindata}^\top X_{\traindata}   + \Sigma_0^{-1}\right)^{-1} \left(\frac{1}{\sigma^2} X_{\traindata}^\top y_{\traindata} + \Sigma_0^{-1}\mu_0\right)\\
    & = \left(\left(X_{\traindata}^\top X_{\traindata}\right)\left(\frac{1}{\sigma^2}I    + \left(X_{\traindata}^\top X_{\traindata}\right)^{-1}\Sigma_0^{-1}\right)\right)^{-1} \left(\frac{1}{\sigma^2} X_{\traindata}^\top y_{\traindata} + \Sigma_0^{-1}\mu_0\right)\\
    & = \left(\frac{1}{\sigma^2}I    + \left(X_{\traindata}^\top X_{\traindata}\right)^{-1}\Sigma_0^{-1}\right)^{-1} \left(X_{\traindata}^\top X_{\traindata}\right)^{-1} \left(\frac{1}{\sigma^2} X_{\traindata}^\top y_{\traindata} + \Sigma_0^{-1}\mu_0\right)\\
    & = \left(\frac{1}{\sigma^2}I    + \left(X_{\traindata}^\top X_{\traindata}\right)^{-1}\Sigma_0^{-1}\right)^{-1} \left(\frac{1}{\sigma^2} \left(X_{\traindata}^\top X_{\traindata}\right)^{-1} X_{\traindata}^\top y_{\traindata} + \left(X_{\traindata}^\top X_{\traindata}\right)^{-1}\Sigma_0^{-1}\mu_0\right)\label{app: eq: post mean}
  \end{align}
  Now, based on the \cref{app: assump: test data}, we have \cref{app: eq: x_x} and \cref{app: eq: x_y}.

  \begin{align}
    X_{\traindata + c \testdata}^\top X_{\traindata + c \testdata} &= \begin{bmatrix}
      X_\traindata\\
      X_\traindata\\
      \vdots\\
      X_\traindata
    \end{bmatrix}^\top
    \begin{bmatrix}
      X_\traindata\\
      X_\traindata\\
      \vdots\\
      X_\traindata
    \end{bmatrix}
    =
    (c + 1)X_{\traindata}^\top X_{\traindata}  \label{app: eq: x_x}
  \end{align}
  \begin{align}
    X_{\traindata + c \testdata}^\top y_{\traindata + c \testdata}
    & = \begin{bmatrix}
      X_\traindata\\
      X_\traindata\\
      \vdots\\
      X_\traindata
    \end{bmatrix}^\top \begin{bmatrix}
      y_\traindata\\
      y_\traindata + \Delta\\
      \vdots\\
      y_\traindata + \Delta
    \end{bmatrix}
    = X_\traindata^\top ((c + 1) y_\traindata + c \Delta) \label{app: eq: x_y}
  \end{align}
  Plugging \cref{app: eq: x_x} and \cref{app: eq: x_y} into \cref{app: eq: post mean} and \cref{app: eq: post cov} we get
  \begin{align}
    \Sigma_{\traindata + c\testdata} & = \left(\left((c + 1)X_{\traindata}^\top X_{\traindata}\right)\left(\frac{1}{\sigma^2}I    + \left((c + 1)X_{\traindata}^\top X_{\traindata}\right)^{-1}\Sigma_0^{-1}\right)\right)^{-1}\\
    & \approx \left(\left((c + 1)X_{\traindata}^\top X_{\traindata}\right)\left(\frac{1}{\sigma^2}I\right)\right)^{-1}\\
    & = \frac{1}{c + 1}\left(\frac{1}{\sigma^2} X_{\traindata}^\top X_{\traindata}\right)^{-1}.
  \end{align}

  \begin{align}
    \mu_{\traindata + c \testdata} & = \left(\frac{1}{\sigma^2}I    + \left((c + 1)X_{\traindata}^\top X_{\traindata}\right)^{-1}\Sigma_0^{-1}\right)^{-1} \nonumber \\
    & \phantom{\text{blah blah}}\left(\frac{1}{\sigma^2} \left((c + 1)X_{\traindata}^\top X_{\traindata}\right)^{-1} X_{\traindata}^\top \left((c + 1)y_{\traindata} + c \Delta\right) + \left((c + 1)X_{\traindata}^\top X_{\traindata}\right)^{-1}\Sigma_0^{-1}\mu_0\right)\\
    & \approx \left(\frac{1}{\sigma^2}I \right)^{-1} \left(\frac{1}{\sigma^2} \left(X_{\traindata}^\top X_{\traindata}\right)^{-1} X_{\traindata}^\top \left(y_{\traindata} + \frac{c}{c + 1}\Delta\right)\right)\\
    & = X_\traindata^{+}(y_\traindata + \frac{c}{c + 1}\Delta).
  \end{align}
  where the $X_\traindata^{+}$ is the Moore-Penrose pseudoinverse of $X_\traindata$ and the $\approx$ follows from \cref{app: assump: vanishing prior}.
\end{proof}

\begin{lem}
  \label{app: lem: lin reg kl divergence of two posterior}
  Let $p$ be the Bayesian linear regression model from \cref{app: defn: linear regression model}.
  Let \Cref{app: assump: test data,app: assump: vanishing prior} hold.
  Let $\alpha$ and $\beta$ be two non-negative integers.
  Then,
  \begin{align}
    \KL{p(z \vert \traindata + \alpha \testdata)}{p(z \vert \traindata + \beta \testdata)}
    \approx \frac{1}{2}\left(k_{\alpha, \beta} d + \Delta^\top M_{\alpha, \beta} \Delta\right), 
  \end{align}
  where $k_{\alpha, \beta}$ is a positive constant and $M_{\alpha, \beta}$ is a positive definite matrix such that
  \begin{align}
    k_{\alpha, \beta} =  \frac{\beta + 1}{\alpha + 1} + \log \frac{\alpha + 1}{\beta + 1} - 1 \quad \quad \text{and} \quad \quad
    M_{\alpha, \beta} 
    = \frac{\left(\beta - \alpha\right)^2}{\left(\alpha + 1\right)^2\left(\beta + 1\right)}\frac{1}{\sigma^2} X_\traindata X_\traindata^{+}.
  \end{align}
\end{lem}
\begin{proof}
  The result follows directly from plugging the approximate mean and the covariance from \cref{app: lem: lin reg posterior with test data} into the expression for KL divergence between the two Gaussians.
  \begin{align}
    \KL{\N(\mu_1, \Sigma_1)}{\N(\mu_2, \Sigma_2)}
    & = \frac{1}{2} \left( \tr(\Sigma_2^{-1} \Sigma_1) + \log \frac{\det \Sigma_2}{\det \Sigma_1} - d + (\mu_2 - \mu_1)^\top \Sigma_2^{-1} (\mu_2 - \mu_1) \right)
  \end{align}
  Collecting the terms apart from the quadratic terms and plugging the covariance expressions from \cref{app: lem: lin reg posterior with test data} for the distributions $p(z \vert \traindata + \alpha \testdata)$ and $p(z \vert \traindata + \beta \testdata)$, we get
  \begin{align}
    \tr(\Sigma_2^{-1} \Sigma_1)  + \log \frac{\det \Sigma_2}{\det \Sigma_1} - d
    & \approx \left(\frac{\beta + 1}{\alpha + 1} + \log \frac{\alpha + 1}{\beta + 1} - 1\right)d
  \end{align}
  And plugging in the expressions for the mean and the covariance from \cref{app: lem: lin reg posterior with test data} for $p(z \vert \traindata + \alpha \testdata)$ and $p(z \vert \traindata + \beta \testdata)$ in the quadratic term, we get
  \begin{align}
    (\mu_2 - \mu_1)^\top \Sigma_2^{-1} (\mu_2 - \mu_1)
    & \approx \frac{\left(\beta - \alpha\right)^2}{\left(\alpha + 1\right)^2\left(\beta + 1\right)}\frac{1}{\sigma^2} \Delta^\top X_\traindata X_\traindata^{+} \Delta
  \end{align}
  Plugging these back into the KL-divergence expression gives the result.
\end{proof}

\begin{thm}[Repeated for convenience]
  \label{app: thm: lin reg snr}
  Let $p(y_\traindata, z)$ be the Bayesian linear regression model where
   $y_\traindata$ be the training response vector. Let $X_\traindata$ be the training feature matrix.
  Let $\traindata_\Delta$ be the mismatched data generated by adding mismatch vector $\Delta$ to $y_\traindata$ such that $y_{\traindata_\Delta} = y_{\traindata} + \Delta$ and $X_{\traindata_\Delta} = X_{\traindata}$.
  Let $\testdata$ be the test data with $m$ copies of $\traindata_{\Delta}$ where $m$ is a positive integer.
  Let $R_K$ be the naive Monte Carlo estimator for $\PPDp$ as in \cref{eq: naive mc estimator for ppdq.}. 
  Then, $\SNR{R_K} =\sqrt{K}/\sqrt{\exp(\delta)^2 - 1}$, where
  \begin{align}
    \lim_{\mathsmaller{\left(X^\top_\traindata X_\traindata\right)^{-1}\Sigma_0^{-1} \to 0}}\, \delta\,
      & =\, \frac{1}{2}d\log \frac{1 + m}{\sqrt{1 + 2m}} \, + \,  \frac{1}{2\sigma^2}\frac{m^2}{2m^2 + 3m + 1}\Delta^\top X_\traindata \left(X_\traindata^\top X_\traindata\right)^{-1} X_\traindata^\top \Delta
      \label{app: eq: lin reg snr}
  \end{align}
 where 
 $d$ is the dimension of feature space.
Furthermore, the following bounds hold
 \begin{align}
  \frac{d}{4}\log \frac{m}{2} \le \lim_{\mathsmaller{\left(X^\top_\traindata X_\traindata\right)^{-1}\Sigma_0^{-1} \to 0}} \, \delta \quad\le \frac{d}{4} \log \left(\frac{m}{2} + 1\right) + \frac{1}{4\sigma^2} ||\Delta||_2^2.
\label{app: eq: lin reg snr bounds}
 \end{align}
\end{thm}

\begin{proof}
  $\delta$ can be written in terms of the KL-divergences between the posteriors $p(z \vert \traindata + \testdata)$ and $p(z \vert \traindata)$ and between the posteriors $p(z \vert \traindata + \testdata)$ and $p(z \vert \traindata + 2\testdata)$.
  From the expressions of KL divergences in \cref{app: lem: lin reg kl divergence of two posterior}, we get 
  \begin{align}
    \delta \approx \frac{1}{4}\left(k d + \Delta^\top M \Delta\right),
  \end{align}
  where
  \begin{align}
    k = k_{m, 0} + k_{m, 2m} \quad \quad \text{and} \quad \quad M = M_{m, 0} + M_{m, 2m}.
  \end{align}
Simplifying the expressions for $k$ and $M$, we get
  \begin{align}
    k 
    &= k_{m, 0} + k_{m, 2m}\\
    &= \log \frac{1 + m}{1} + \frac{1}{1 + m} - 1 + \log \frac{1 + m }{1 + 2m} + \frac{1 + 2m}{1 + m} - 1\\
    &= \log \frac{1 + m}{1} + \log \frac{1 + m }{1 + 2m}\\
    &= \log \frac{\left(1 + m\right)^2 }{1 + 2m}\\
    &= 2\log \frac{1 + m}{\sqrt{1 + 2m}}\\
  \end{align}
  and
  \begin{align}
    M 
    &= M_{m, 0} + M_{m, 2m}\\
    &= \frac{m^2}{(m+1)^2}\frac{1}{\sigma^2} X_\traindata X_\traindata^{+} +  \frac{m^2}{\left(m+1\right)^2\left(2m+1\right)} \frac{1}{\sigma^2} X_\traindata X_\traindata^{+} \\
    &= \frac{m^2}{(m+1)^2}\frac{1}{\sigma^2} X_\traindata X_\traindata^{+} \left(1 + \frac{1}{2m+1}\right) \\
    &= \frac{m^2}{(m+1)^2} \frac{2\left(m + 1\right)}{2m+1}\frac{1}{\sigma^2} X_\traindata X_\traindata^{+}  \\
    &= \frac{2m^2}{\left(m+1\right) \left(2m+1\right)} \frac{1}{\sigma^2} X_\traindata X_\traindata^{+}  \\
  \end{align}
\end{proof}

The main assumption in \cref{app: thm: lin reg snr} is $\left(X_\traindata^\top X_\traindata\right)^{-1} \Sigma_0^{-1} \to 0$. This essentially means that the feature matrix $X_\traindata$ and prior parameters $\left(\mu_0, \Sigma_0\right)$ are such that the posterior parameters $(\mu_\traindata, \Sigma_\traindata)$ are not influenced by the prior. This is analogous to the assumption made in \cref{prop: delta Bayesian CLT} where we assume that the training and test datasets are big enough such that Bayesian CLT holds (Note that for the case where there is no mismatch, that is $\Delta = 0$, the expression for the limit in \cref{eq: lin reg snr} reduces to the $\delta$ approximation in  \cref{eq: delta CLT}).

Moreover, the limit in \cref{app: eq: lin reg snr} can be bounded by bounding three individual terms. First, $\log (1+m)/\sqrt{1 + 2m}$ is lower-bounded by $d/4 \log (m/2)$ and upper-bounded by $d/4 \log (m/2 + 1)$. Second, $m^2/\left(2m^2 + 3m + 1\right)$ is lower-bounded by $1/6$ and upper-bounded by $1/2$. Third, we have 
\begin{align}
  \Delta^\top X_\traindata \left( X_\traindata^\top X_\traindata\right)^{-1} X_\traindata^\top \Delta = \Delta^\top UU^\top \Delta
\end{align}
where $U$ is the left singular matrix of $X_\traindata$ containing $d$ singular left vectors. Then, from the properties of the left-singular vectors, $||U^\top \Delta||_2^2$ terms is lower-bounded by $0$ and upper-bounded by $||\Delta||_2^2$. Combining these bounds, we get the bounds in \cref{app: eq: lin reg snr bounds}.

  Overall, \Cref{app: thm: lin reg snr} captures the strength of three factors that affect the SNR of the naive MC estimator: (i) the mismatch between train and test data\textemdash$\delta$ scales quadratically in $\Delta$, (ii) the dimensionality of the latent variable\textemdash$\delta$ scales linearly in $d$, and (iii) the ratio of the size of test data and training data\textemdash$\delta$ scales logarithmically in $m$. 

  \subsection{Experimental Details}
  \label{app: sec: linear regression example experiment details}
  We consider the linear regression model with likelihood $p(y_\traindata \vert z) = \mathcal{N}(y_\traindata \vert X_\traindata z, \sigma^2 I).$ 
  where $y_\traindata \in \R^{\verts{\traindata}}$ is the response vector, $X_\traindata \in \R^{\verts{\traindata}\times d}$ is feature matrix, and $\sigma^2$ is the variance. 
  The conjugate prior is 
  $p(z) = \mathcal{N}(z \vert \mu_0, \Sigma_0)$
  where $z \in \R^d$ $\mu_0$ is the mean and $\Sigma_0$ is the covariance. 
  
  We consider the exact inference settings and start with a baseline scenario where none of the three factors influencing SNR are too high. 
  Thereafter, we independently increase the three factors: mismatch, the dimensionality of the latent space, and the size of the test data to create three additional scenarios. We use the standard normal prior and likelihood with $\sigma^2 = 1$.
  
  \textbf{Baseline.} We set the number of training data points to $1000$, the dimensionality of laten space $d = 10$, and the number of mismatched copies $m = 1$. We then forward sample a training data set $\traindata$ and then generate the mismatched data $\traindata_\Delta$ by adding a mismatch vector $\Delta = 2$ to the response vector $y_\traindata$. 
  
  \textbf{More mismatch.} We keep the training data same as in the baseline scenario and increase the mismatch vector to $\Delta = 10$.
  
  \textbf{More test data.} We keep the training data same as in the baseline scenario and increase the number of mismatched copies to $m = 10$.
  
  \textbf{More dimensions.} We keep the number of training data points, the number of mismatched copies, and the mismatch vector same as in the baseline scenario and increase the dimensionality of the latent space to $d = 100$. We forward sample the training data set $\traindata$ and then generate the mismatched data $\traindata_\Delta$ by adding a mismatch vector $\Delta = 2$ to the response vector $y_\traindata$.
  
  \Cref{fig: lin reg estimation error} reports the results from estimating $\PPDp$ using naive MC estimator $R_K$ from \cref{eq: naive mc estimator for ppdq.} for $K = 10^0, 10^1, \dots, 10^6$.
  The error bands are the $95\%$ confidence intervals based on $1000$ independent evaluations.

  For LIS, we learn a full-rank Gaussian proposal distribution by optimizing the IW-ELBO from \cref{eq: iwelbo under q} with $M = 16$ using the DReG estimator and ADAM optimizer with a learning rate of $0.001$ for $1000$ iterations.
  We consider different initialization techniques for the variational parameters: Laplace's approximation and standard Normal, and pick the one that provides higher initial ELBO. For each optimization step, we use 8 copies to average the IW-ELBO gradient.
  For LIS, we learn the proposal once, and do $1,000$ independent evaluations to estimate the error bands.


\newpage

\section{Logistic Regression: Additional Details}
\label{app: final sec: logistic regression.}

We consider the logistic regression model with likelihood $p(y \vert z) = \mathcal{B}(\mathrm{sigmoid}(x^\top z))$ where $y \in \{0,1\}$ is the binary response, $x \in \R^d$ is the feature vector, $z \in \R^d$ is the latent variable, and $\mathcal{B}$ is the Bernoulli distribution. The non-conjugate prior $p(z)$ is given by a normal distribution $\mathcal{N}(z \vert \mu_0, \Sigma_0)$.
We set the prior to standard Normal for the experiments.

We structure our experiments in a similar way as the linear regression model. Here the mismatch between the training and test data is created by flipping the first $\Delta$ fraction of the response vector $y_\traindata$ to create the mismatched data $\traindata_\Delta$. 

\textbf{Baseline.} We set the number of training data points to $1000$, the dimensionality of latent space $d = 10$, and the number of mismatched copies $m = 1$. We forward sample a training data set $\traindata$ and then generate the mismatched data $\traindata_\Delta$ by adding flipping the first $\Delta = 0.1$ fraction of the response vector $y_\traindata$.

\textbf{More mismatch.} We keep the training data same as in the baseline scenario and increase the mismatch fraction to $\Delta = 1.0$.

\textbf{More test data.} We keep the training data same as in the baseline scenario and increase the number of mismatched copies to $m = 10$.

\textbf{More dimensions.} We keep the number of training data points, the number of mismatched copies, and the mismatch fraction same as in the baseline scenario and increase the dimensionality of the latent space to $d = 100$. We forward sample the training data set $\traindata$ and then generate the mismatched data $\traindata_\Delta$ by flipping the first $\Delta = 0.1$ fraction of the response vector $y_\traindata$.

We learn a full-rank Gaussian variational approximation by optimizing the standard ELBO objective using the ADAM optimizer with a learning rate of $0.001$ for $1000$ iterations. We consider different initialization techniques for the variational parameters: Laplace's approximation and standard Normal, and pick the one that provides higher initial ELBO. For each optimization step, we use 16 independent copies to average the ELBO gradient.

For LIS, we learn a full-rank Gaussian proposal distribution by optimizing the IW-ELBO from \cref{eq: iwelbo under q} with $M = 16$ using the ADAM optimizer with a learning rate of $0.001$ for $1000$ iterations. We consider different initialization techniques for the variational parameters: Laplace's approximation and standard Normal, and pick the one that provides higher initial ELBO. We use a 8 copies to average the gradient of IW-ELBO.

\newpage

\section{Hierarchical Model: Additional Details}
\label{app: final sec: hierarchical model.}
We use MovieLens25M \cite{harper2015movielens}, a dataset of 25 million movie ratings with over 60,000 movies, rated by more than 160,000 users. We also use set of features for each movie (tag relevance scores \cite{vig2012tag}.)

Movielens25M originally uses a 5 point ratings system. To get binary ratings, we map ratings greater than 3 points to 1 and less than and equal to 3 to 0.
We pre-process the data to drop users with more than 1,000 ratings---leaving around $20$M ratings.
Also, we PCA the movie features to reduce their dimensionality to 10. 
We used a train-test split such that, for each user, one-tenth of the ratings are in the test set. 
This gives us $\approx$ 18M ratings for training (and $\approx$ 2M ratings for testing.) Our of these we randomly select $100$ users for experiments. 

For Gaussian VI, we use a full-rank Gaussian. We optimize standard ELBO using ADAM for $1000$ iterations with step-size of $0.001$. For each optimization step, we use 16 copies to average the gradient.


For flow VI, we use a  real-NVP flow with 10 coupling layers for all our experiments. We define each coupling layer to be comprised of two transitions, where a single transition corresponds to affine  transformation of one part of the latent variables. For example, if the input variable for the $k^{th}$ layer is $z^{(k)}$, then first transition is defined as     
\begin{align}  
  z_{1:d} &= z^{(k)}_{1:d}\nonumber\\
  z_{d+1:D} &= z^{(k)}_{d+1:D} \odot \exp\big(s^{a}_{k}(z^{(k)}_{1:d})\big) + t^{a}_{k}(z^{(k)}_{1:d})).
  \label{eq:rnvp-appendix}
\end{align}  
where, for the function $s$ and $t$, super-script $a$ denotes first transition and sub-script $k$ denotes the $k^{th}$ layer. For the next transition, the $z_{d+1:D}$ part is kept unchanged and $z_{1:d}$ is affine transformed in a similar fashion to obtain the layer output $z^{(k+1)}$ (this time using $s^{b}_{k} (z^{(k)}_{d+1:D}) $ and $t^{b}_{k} (z^{(k)}_{d+1:D}) $ ). 
This is also referred to as the alternating first half binary mask. 
Both, scale($s$) and translation($t$) functions of single transition are parameterized by the same fully connected neural network(FNN). More specifically, for first transition in above example, a single FNN takes $z^{(k)}_{1:d}$ as input and outputs both $s^{a}_{k}(z^{(k)}_{1:d})$ and $t^{a}_{k}(z^{(k)}_{1:d})$. Thus, the skeleton of the FNN, in terms of the size of the layers, is as $[d, H,H, 2(D-d)]$ where, $H$ denotes the size of the two hidden layers ($H$=32 for all our experiments).  

The hidden layers of FNN use a leaky rectified linear unit with slope = 0.01, while the output layer uses a hyperbolic tangent for $s$ and remains linear for $t$. 
We initialize the parameters of the neural networks from normal distribution $\mathcal{N}(0, 0.001^{2})$. This choice approximates standard normal initialization.
We optimize standard ELBO with sticking the landing (STL) \citep{roeder2017sticking} gradient using ADAM for $1000$ iterations with step-size of $0.001$. For each optimization step, we use 16 copies to average the gradient.


To learn the proposal distribution for the learn IS estimator, we use a realNVP flow with architecture described above. We initialize it with parameters from the variational distribution. For the Gaussian VI, we fix the base distribution for the flow to the variational distribution. For flow VI, we use the same architecture for the proposal distribution and simply initialize using the parameters of the variational distribution.
We optimize IW-ELBO with DReG estimator  using ADAM for $100$ iterations with step-size of $0.001$. For each optimization step, we use $8$ copies to average the gradient.
\newpage

\end{document}

%% file: results/ppd_df.tex
\begin{tabular}{@{}lrrrrr@{}}
\toprule
Model & $\log \PPDp$ & $\E [\log R_K]$ & $\E [\log R_K^{\mathrm{IS}}]$ & $\textrm{SNR}(R_K)$ & $\textrm{SNR}(R_K^{\mathrm{IS}})$ \\
\midrule
Normal & -774.64 & -1183.98 $\pm$ 0.34 & -774.64 $\pm$ 0.00 & 0.35 $\pm$ 0.02 & 76.5 $\pm$ 23.43 \\
Exp & -527.44 & -559.98 $\pm$ 0.16 & -527.44 $\pm$ 0.00 & 0.34 $\pm$ 0.01 & 222.76 $\pm$ 147.59 \\
Binomial & -327.42 & -487.13 $\pm$ 1.29 & -327.41 $\pm$ 0.00 & 0.03 $\pm$ 0.00 & 173.06 $\pm$ 104.2 \\
\bottomrule
\end{tabular}

%% file: results/approx_ppd_df.tex
\begin{tabular}{@{}lcrrrr@{}}
\toprule
Model & $\log \PPDp$ & $\E[{\log R_{K}}]$ & $\E[{\log R_{K}^{\mathrm{IS}}}]$ & ${\textrm{SNR} (R_{K})}$ & ${\textrm{SNR} (R_{K}^{\mathrm{IS}})}$ \\
\midrule
Normal & -- & -1194.32 $\pm$ 0.41 & -775.23 $\pm$ 0.00 & 0.35 $\pm$ 0.02 & 238.79 $\pm$ 172.46 \\
Exp & -- & -576.27 $\pm$ 0.14 & -542.34 $\pm$ 0.00 & 0.36 $\pm$ 0.01 & 215.09 $\pm$ 140.52 \\
Binomial & -- & -382.46 $\pm$ 0.74 & -322.66 $\pm$ 0.00 & 0.34 $\pm$ 0.02 & 70.13 $\pm$ 35.29 \\
\bottomrule
\end{tabular}

%% file: results/log_reg_ppdq.tex
\begin{tabular}{@{}lcrrrr@{}}
\toprule
 & $\log \mathrm{PPD}$ & $\E[\log R_{K}]$ & $\E[\log R_{K}^{\mathrm{IS}}]$ & $\mathrm{SNR}(R_{K})$ & $\mathrm{SNR}(R_{K}^{\mathrm{IS}})$ \\
\midrule
Baseline & - & -525.25 $\pm$ 0.01 & -525.12 $\pm$ 0.00 & 1.35 $\pm$ 0.20 & 645.67 $\pm$ 14.99 \\
More dimension & - & -702.07 $\pm$ 0.28 & -543.00 $\pm$ 0.00 & 0.04 $\pm$ 0.01 & 57.78 $\pm$ 1.37 \\
More mismatch & - & -1687.98 $\pm$ 0.96 & -734.32 $\pm$ 0.00 & 0.03 $\pm$ 0.00 & 728.63 $\pm$ 16.01 \\
More test data & - & -5143.69 $\pm$ 0.34 & -5097.60 $\pm$ 0.00 & 0.04 $\pm$ 0.01 & 802.34 $\pm$ 18.37 \\
\bottomrule
\end{tabular}

%% file: results/movielens_results.tex
\begin{tabular}{@{}lcrrrrrrr@{}}
\toprule
 & \multicolumn{2}{c}{$\E[\log R_K]$} & \multicolumn{2}{c}{$\E[\log R_K^{\mathrm{IS}}]$} & \multicolumn{2}{c}{$\mathrm{SNR}(R_K)$} & \multicolumn{2}{c}{$\mathrm{SNR}(R_K^{\mathrm{IS}})$} \\
 & $K = 10^{3}$ & $K = 10^{6}$ & $K = 10^{3}$ & $K = 10^{6}$ & $K = 10^{3}$ & $K = 10^{6}$ & $K = 10^{3}$ & $K = 10^{6}$ \\
\midrule
Flow VI & -796.24 $\pm$ 0.13 & -787.27 $\pm$ 0.08 & -779.39 $\pm$ 0.02 & -777.73 $\pm$ 0.01 & 0.05 $\pm$ 0.02 & 0.04 $\pm$ 0.01 & 0.11 $\pm$ 0.04 & 0.48 $\pm$ 0.29 \\
Gaussian VI & -828.22 $\pm$ 0.17 & -811.61 $\pm$ 0.13 & -783.89 $\pm$ 0.03 & -781.88 $\pm$ 0.02 & 0.04 $\pm$ 0.00 & 0.04 $\pm$ 0.01 & 0.12 $\pm$ 0.01 & 0.32 $\pm$ 0.13 \\
\bottomrule
\end{tabular}